\documentclass{article}

\usepackage{times}
\usepackage{graphicx} % more modern
\usepackage{subfigure} 

\usepackage{natbib}

\usepackage{algorithm}
\usepackage{algorithmic}
\usepackage{bbm}

\usepackage[arxiv]{arxiv_icml2017} 

\icmltitlerunning{An Analytic Formula of Population Gradient for 2-layered ReLU and its Applications}

\usepackage{amsmath,amsfonts,amssymb}
\usepackage{graphicx}
\usepackage{url}

\newcounter{ThmCounter}
\newcounter{PropCounter}
\newcounter{ConjectureCounter}

\newtheorem{theorem}[ThmCounter]{Theorem}
\newtheorem{proposition}[PropCounter]{Proposition}
\newtheorem{conjecture}[ConjectureCounter]{Conjecture}

\newcommand{\qed}{\nobreak \ifvmode \relax \else
    \ifdim\lastskip<1.5em \hskip-\lastskip
    \hskip1.5em plus0em minus0.5em \fi \nobreak
\vrule height0.75em width0.5em depth0.25em\fi}

\def\dd{\mathrm{d}}
\def\vw{\mathbf{w}}
\def\vm{\mathbf{m}}

\def\vx{\mathbf{x}}

\def\vu{\mathbf{u}}
\def\vv{\mathbf{v}}

\def\ve{\mathbf{e}}
\def\opt{^{*}}

\def\t{^{t}}
\def\tone{^{0}}
\def\tnext{^{t+1}}
\def\trans{^\intercal}
\def\ee#1{\mathbb{E}\left[#1\right]}
\def\eex#1{\mathbb{E}_X\left[#1\right]}

\def\dd{\mathrm{d}}

\def\vone{\mathbf{1}}

\def\lenw{\|\vw\|}

\def\diag{\mathrm{diag}}

\def\newcite#1{\mbox{\cite{#1}}}

\begin{document} 

\twocolumn[
    \icmltitle{An Analytical Formula of Population Gradient for two-layered ReLU network and its Applications in Convergence and Critical Point Analysis}

% It is OKAY to include author information, even for blind
% submissions: the style file will automatically remove it for you
% unless you've provided the [accepted] option to the icml2017
% package.

% list of affiliations. the first argument should be a (short)
% identifier you will use later to specify author affiliations
% Academic affiliations should list Department, University, City, Region, Country
% Industry affiliations should list Company, City, Region, Country

% you can specify symbols, otherwise they are numbered in order
% ideally, you should not use this facility. affiliations will be numbered
% in order of appearance and this is the preferred way.
\icmlsetsymbol{equal}{*}

\begin{icmlauthorlist}
    \icmlauthor{Yuandong Tian}{fair}
\end{icmlauthorlist}

\icmlaffiliation{fair}{Facebook AI Research}
\icmlcorrespondingauthor{Yuandong Tian}{yuandong@fb.com}

% You may provide any keywords that you 
% find helpful for describing your paper; these are used to populate 
% the "keywords" metadata in the PDF but will not be shown in the document
\icmlkeywords{ICML, deep learning, nonconvex optimization, convergence analysis}

\vskip 0.3in
]

% this must go after the closing bracket ] following \twocolumn[ ...

% This command actually creates the footnote in the first column
% listing the affiliations and the copyright notice.
% The command takes one argument, which is text to display at the start of the footnote.
% The \icmlEqualContribution command is standard text for equal contribution.
% Remove it (just {}) if you do not need this facility.

\printAffiliationsAndNotice  % leave blank if no need to mention equal contribution
%\printAffiliationsAndNotice{\icmlEqualContribution} % otherwise use the standard text.
%\footnotetext{hi}

\begin{abstract}
    In this paper, we explore theoretical properties of training a two-layered ReLU network $g(\vx; \vw) = \sum_{j=1}^K \sigma(\vw_j\trans\vx)$ with centered $d$-dimensional spherical Gaussian input $\vx$ ($\sigma$=ReLU). We train our network with gradient descent on $\vw$ to mimic the output of a teacher network with the same architecture and fixed parameters $\vw\opt$. We show that its population gradient has an analytical formula, leading to interesting theoretical analysis of critical points and convergence behaviors. First, we prove that critical points outside the hyperplane spanned by the teacher parameters (``out-of-plane``) are not isolated and form manifolds, and characterize in-plane critical-point-free regions for two ReLU case. On the other hand, convergence to $\vw\opt$ for one ReLU node is guaranteed with at least $(1-\epsilon)/2$ probability, if weights are initialized randomly with standard deviation upper-bounded by $O(\epsilon/\sqrt{d})$, consistent with empirical practice. For network with many ReLU nodes, we prove that an infinitesimal perturbation of weight initialization results in convergence towards $\vw\opt$ (or its permutation), a phenomenon known as spontaneous symmetric-breaking (SSB) in physics. We assume no independence of ReLU activations. Simulation verifies our findings. 
\end{abstract}

\section{Introduction}
Despite empirical success of deep learning (e.g., Computer Vision~\cite{resnet,vgg,inception,alexnet}, Natural Language Processing~\newcite{sutskever2014sequence} and Speech Recognition~\newcite{hinton2012deep}), it remains elusive how and why simple methods like gradient descent can solve the complicated non-convex optimization in its training procedure. In this paper, we focus on the following two-layered ReLU network:
\begin{equation}
    g(\vx; \vw) = \sum_{j=1}^K \sigma(\vw_j\trans\vx), \label{eq:network-formulation}
\end{equation}
Here $\sigma(x) = \max(x, 0)$ is the ReLU nonlinearity. We consider the setting that a student network is optimized to minimize the $l_2$ distance between its prediction and the supervision provided by a teacher network of the same architecture with fixed parameters $\vw\opt$. Note that although the network prediction (Eqn.~\ref{eq:network-formulation}) is convex due to the property of ReLU, if coupled with loss functions (e.g., $l_2$ loss Eqn.~\ref{eq:l2loss}), then the optimization becomes highly non-convex and has exponential number of critical points.

To analyze it, we introduce a simple analytic formula for population gradient in the case of $l_2$ loss, when inputs $\vx$ are sampled from zero-mean spherical Gaussian. Using this formula, critical point and convergence analysis follow. 

For critical points, we show that critical points outside the principal hyperplane (the subspace spanned by $\vw\opt$) form manifolds. We also characterize the region in the principal hyperplane that has no critical points, in two ReLU case.  

We also analyze the convergence behavior under the population gradient. Using Lyapunov method~\cite{lyapunov}, for single ReLU case we prove that gradient descent converges to $\vw\opt$ with at least $(1-\epsilon)/2$ probability, if initialized randomly with standard deviation upper-bounded by $O(\epsilon/\sqrt{d})$, verifying common initialization techniques~\cite{bottou-88b, xavier, PReLU, lecun2012efficient},. For multiple ReLU case, when the teacher parameters $\{\vw_j\}_{j=1}^K$ form a set of orthonormal basis, we prove that \textbf{(1)} a symmetric weight initialization gets stuck at a saddle point and \textbf{(2)} a particular infinitesimal perturbation of (1) leads to convergence towards $\vw\opt$ or its permutation. This behavior is known as spontaneous symmetry breaking in physics, in which the population gradient field enjoys invariance under a certain kind of symmetry, but the solution breaks it. Although such behaviors have been known empirically, to our knowledge, this paper first formally characterizes them in 2-layered ReLU network.

\begin{figure}[t]
    \centering
    \includegraphics[width=0.5\textwidth]{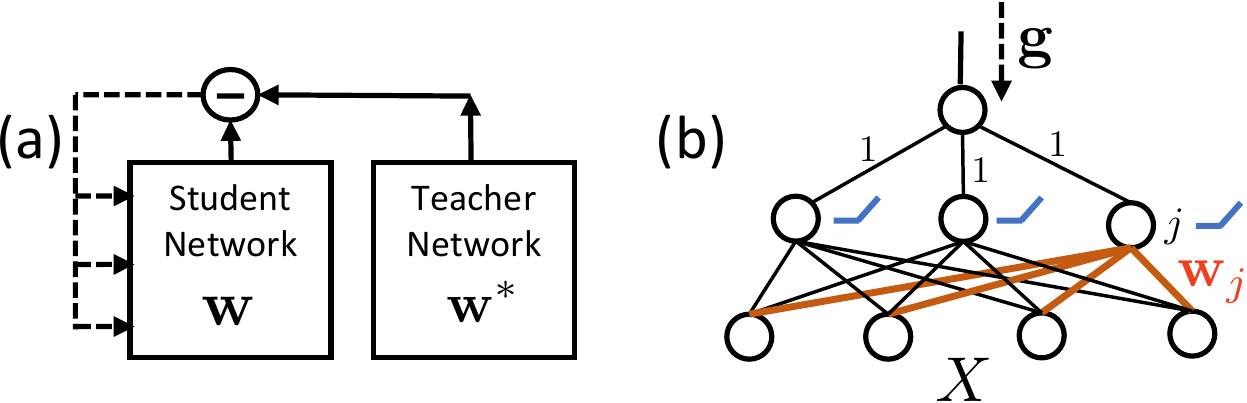}
    \caption{\textbf{(a)} We consider the student and teacher network as nonlinear neural networks with ReLU nonlinearity. The student network updates its weight $\vw$ from the output of the teacher with fixed weights $\vw\opt$. \textbf{(b)} The 2-layered ReLU network structure (Eqn.~\ref{eq:network-formulation}) discussed in this paper. The first layer contains fixed weights of value $1$, while the second layers has $K$ ReLU nodes. Each node $j$ has a $d$-dimensional weight $\vw_j$ to be optimized. Teacher network has the same architecture as the student.} \label{fig:models}
\end{figure}

\section{Related Works}
For multilayer linear network, many works analyze its critical points and convergence behaviors. \newcite{saxe2013exact} analyzes its dynamics of gradient descent and~\newcite{kawaguchi2016deep} shows every local minimum is global. On the other hand, very few theoretical works have been done for \emph{nonlinear} networks. \newcite{mei2016landscape} shows the global convergence for a single nonlinear node whose derivatives of activation $\sigma'$, $\sigma''$, $\sigma'''$ are bounded and $\sigma'> 0$. Similar to our approach, \newcite{saad1996dynamics} also uses the student-teacher setting and analyzes the student dynamics when the teacher's parameters $\vw\opt$ are orthonormal. However, their activation is Gaussian error function $\mathrm{erf}(x)$, and only the local behaviors of the two critical points (the initial saddle point near the origin and $\vw\opt$) are analyzed. Recent paper~\newcite{eletron-proton} analyzes a similar teacher-student setting on 2-layered network when the involved function is harmonic, but it is unclear how the conclusion is generalized to ReLU case. To our knowledge, our close-form formula for 2-layered ReLU network is novel, as well as the critical point and convergence analysis.  

Many previous works analyze nonlinear network based on the assumption of \emph{independent activations}: the activations of ReLU (or other nonlinear) nodes are independent of the input and/or mutually independent. For example, \newcite{choromanska2015loss,choromanska2015open} relates the nonlinear ReLU network with spin-glass models when several assumptions hold, including the assumption of independent activations (\textbf{A1p} and \textbf{A5u}). \newcite{kawaguchi2016deep} proves that every local minimum in nonlinear network is global based on similar assumptions. \newcite{soudry2016no} shows the global optimality of the local minimum in a two-layered ReLU network, when independent multiplicative Bernoulli noise is applied to the activations. In practice, activations that share the input are highly dependent. Ignoring such dependency misses important behaviors, and may lead to misleading conclusions. In this paper, \emph{no assumption of independent activations is made}. Instead, we assume input to follow spherical Gaussian distribution, which gives more realistic and interdependent activations during training.  

For sigmoid activation, \newcite{fukumizu2000local} gives complicated conditions for a local minimum to be global when adding a new node to a 2-layered network. \newcite{janzamin2015beating} gives guarantees for parameter recovery of a 2-layered network learnt with tensor decomposition. In comparison, we analyze ReLU networks trained with gradient descent, which is more popular in practice. 

\def\grad{\ee{\nabla J}}

\section{Problem Definition}
\label{sec:problem-definition}
Denote $N$ as the number of samples and $d$ as the input dimension. The $N$-by-$d$ matrix $X$ is the input data and $\vw\opt$ is the fixed parameter of the teacher network. Given the current estimation $\vw$, we have the following $l_2$ loss:
\begin{equation}
    J(\vw) = \frac12 \| g(X; \vw\opt) - g(X; \vw) \|^2, \label{eq:l2loss}
\end{equation}
Here we focus on \emph{population loss} $\eex{J}$, where the input $X$ is assumed to follow spherical Gaussian distribution $\mathcal{N}(0, I)$. Its gradient is the \emph{population gradient} $\eex{\nabla J_\vw(\vw)}$ (abbrev. $\grad$). In this paper, we study critical points $\grad = 0$ and vanilla gradient dynamics $\vw\tnext = \vw\t - \eta\ee{\nabla J(\vw\t)}$, where $\eta$ is the learning rate. 

\section{The Analytical Formula}
\label{sec:single-relu-case}
\textbf{Properties of ReLU}. ReLU nonlinearity has useful properties. We define the \emph{gating function} $D(\vw) \equiv \mathrm{diag}(X\vw > 0)$ as an $N$-by-$N$ binary diagonal matrix. Its $l$-th diagonal element is a binary variable showing whether the neuron is activated for sample $l$. Using this notation, $\sigma(X\vw) = D(\vw)X\vw$ which means $D(\vw)$ selects the output of a linear neuron, based on their activations. Note that $D(\vw)$ only depends on the direction of $\vw$ but not its magnitude. 

$D(\vw)$ is also ``transparent'' with respect to derivatives. For example, at differentiable regions, $\mathrm{Jacobian}_\vw[\sigma(X\vw)] = \sigma'(X\vw) X = D(\vw)X$. This gives a very concise rule for gradient descent update in ReLU networks.

\textbf{One ReLU node}. Given the properties of ReLU, the population gradient $\grad$ can be written as:
\begin{equation}
    \grad = \eex{X\trans D(\vw) \left(D(\vw)X\vw - D(\vw\opt) X\vw\opt\right)}  \label{eq:basic-gradient} 
\end{equation}
Intuitively, this term vanishes when $\vw \rightarrow \vw\opt$, and should be around $\frac{N}{2} (\vw - \vw\opt)$ if the data are evenly distributed, since roughly half of the samples are blocked. However, such an estimation fails to capture the nonlinear behavior.

If we define \emph{Population Gating} (PG) function $F(\ve, \vw) \equiv X\trans D(\ve) D(\vw) X\vw$, then $\grad$ can be written as:
\begin{equation}
    \grad = \ee{F(\vw/\lenw, \vw)} - \ee{F(\vw/\lenw, \vw\opt)}.
\end{equation}

\begin{figure}
    \centering
    %\vspace{-0.1in}
    \includegraphics[width=0.45\textwidth]{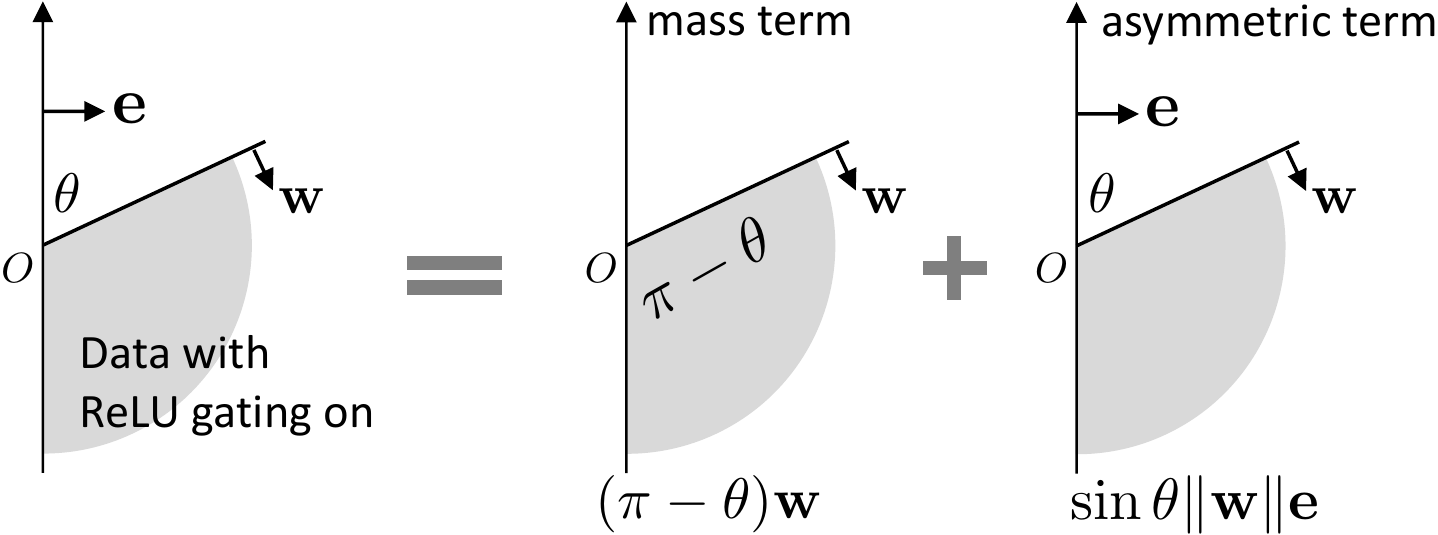}
    \vspace{-0.1in}
    \caption{Decomposition of \emph{Population Gating} (PG) function $F(\ve, \vw)$ (Eqn.~\ref{eq:f}) into mass term and asymmetric term. $F(\ve, \vw)$ is computed from the portion of data with ReLU gate on. The mass term is proportional to the amount of data, while the asymmetric term is related to the data asymmetry with respect to $\ve$.}
    \label{fig:close-form-decomp}
\end{figure}

Interestingly, $F(\ve, \vw)$ has an analytic formula if the data $X$ follow spherical Gaussian distribution:
\begin{theorem}
    Denote $F(\ve, \vw) = X\trans D(\ve) D(\vw) X\vw$ where $\ve$ is a unit vector, $X = [\vx_1, \vx_2, \cdots, \vx_N]\trans$ is the $N$-by-$d$ data matrix and $D(\vw) = \mathrm{diag}(X\vw > 0)$ is a binary diagonal matrix. If $\vx_i \sim \mathcal{N}(0, I)$ (and thus bias-free), then:
    \begin{equation}
        \ee{F(\ve, \vw)} = \frac{N}{2\pi} \left[ (\pi - \theta)\vw + \lenw\sin\theta\ve\right] \label{eq:f}
    \end{equation}
    where $\theta = \angle(\ve, \vw) \in [0, \pi]$ is the angle between $\ve$ and $\vw$. 
\end{theorem}
See the link\footnote{http://yuandong-tian.com/ssb-supp.pdf} for the proof of all theorems. Note that we do not require $X$ to be independent between samples. Intuitively, the first \emph{mass} term $\frac{N}{2\pi}(\pi-\theta)\vw$ aligns with $\vw$ and is proportional to the amount of \emph{activated} data whose ReLU are on. When $\theta = 0$, the gating function is fully on and half of the data contribute to the term; when $\theta = \pi$, the gating function is completely switched off. The gate is controlled by the angle between $\vw $ and the \emph{control} signal $\ve$. The second \emph{asymmetric} term is aligned with $\ve$, and is proportional to the asymmetry of the activated data samples with respect to $\ve$ (Fig.~\ref{fig:close-form-decomp}). 

Note that the expectation analysis smooths out ReLU and leaves only one singularity at the origin, where $\grad$ is not continuous. That is, if approaching from different directions towards $\vw = 0$, $\grad$ is different. 

With the close form of $F$, $\grad$ also has a close form:
\begin{equation}
    \grad = \frac{N}{2} (\vw - \vw\opt) + \frac{N}{2\pi}\left(\theta\vw\opt - \frac{\|\vw\opt\|}{\|\vw\|}\sin\theta\vw \right) \label{eq:dynamics}
\end{equation}
where $\theta =\angle(\vw,\vw\opt) \in [0, \pi]$. The first term is from linear approximation, while the second term shows the nonlinear behavior.

For linear case, $D \equiv I$ (no gating) and thus $\nabla J \propto X\trans X(\vw - \vw\opt)$. For spherical Gaussian input $X$, $\eex{X\trans X} = I$ and $\grad \propto \vw - \vw\opt$. Therefore, the dynamics has only one critical point and global convergence follows, which is consistent with its convex nature. 

\textbf{Extension to other distributions}. From its definition, $\ee{F(\ve,\vw)} = \ee{X\trans D(\ve) D(\vw) X\vw}$ is linear to $\|\vw\|$, regardless of the distribution of $X$. On the other hand, isotropy in spherical Gaussian distribution leads to the fact that $\ee{F(\ve,\vw)}$ only depends on angles between vectors. For other isotropic distributions, we could similarly derive:
\begin{equation}
    \ee{F(\ve, \vw)} = A(\theta)\vw + \lenw B(\theta)\ve \label{eq:f-generalized}
\end{equation}
where $A(0) = N/2$ (gating fully on), $A(\pi) = 0$ (gating fully off), and $B(0) = B(\pi)= 0$ (no asymmetry when $\vw$ and $\ve$ are aligned). Although we focus on spherical Gaussian case, many following analysis, in particular critical point analysis, can also be applied to Eqn.~\ref{eq:f-generalized}.

\def\gradj{\ee{\nabla_{\vw_j} J}} 
\def\gradone{\ee{\nabla_{\vw_1} J}} 

\textbf{Multiple ReLU node.} For Eqn.~\ref{eq:network-formulation} that contains $K$ ReLU node, we could similarly write down the population gradient with respect to $\vw_j$ (note that $\ve_j = \vw_j / \|\vw_j\|$):
\begin{equation}
    \ee{\nabla_{\vw_j} J} = \sum_{j'=1}^K \ee{F(\ve_j, \vw_{j'})} - \sum_{j'=1}^K \ee{F(\ve_j, \vw\opt_{j'})} \label{eq:two-layer-fix-top} 
\end{equation}

\section{Critical Point Analysis}
\def\normvw{\bar\vw}
\def\normvwopt{\bar\vw\opt}
By solving Eqn.~\ref{eq:two-layer-fix-top} (the \emph{normal equation}, $\gradj = 0$), we could identify all critical points of $g(\vx)$. However, it is highly nonlinear and cannot be solved easily. In this paper, we provide conditions for critical points using the structure of Eqn.~\ref{eq:two-layer-fix-top}. Following the analysis, the case study for $K=2$ gives examples for saddle points and regions without critical points. 

For convenience, we define $\Pi_*$ as the \emph{Principal Hyperplane} spanned by $K$ ground truth weight vectors. Note that $\Pi_*$ is at most $K$ dimensional. $\{\vw_j\}_{j=1}^K$ is said to be \emph{in-plane}, if all $\vw_j \in \Pi_*$. Otherwise it is \emph{out-of-plane}.

\subsection{Normal Equation}
\def\normvw{\bar\vw}
\def\normvwopt{\bar\vw\opt}
The normal equation $\{\gradj = 0\}_{j=1}^K$ contain $Kd$ scalar equations and can be written as the following:
\begin{eqnarray}
    Y E\trans = B\opt{W\opt}\trans \label{eq:normal-full}
\end{eqnarray}
where $Y = \diag(\sin\Theta\trans \normvw - \sin{\Theta\opt}\trans \normvwopt) + (\pi\vone\vone\trans - \Theta\trans)\diag\normvw$ and $B\opt = \pi\vone\vone\trans - (\Theta\opt)\trans$. Here $\theta_j^{*j'} \equiv \angle (\vw_j, \vw\opt_{j'})$, $\theta_j^{j'} \equiv \angle (\vw_j,\vw_{j'})$, $\Theta = [\theta^i_j]$ ($i$-th row, $j$-th column of $\Theta$ is $\theta^i_j$) and $\Theta\opt = [\theta^{*i}_j]$.  

Note that $Y$ and $B\opt$ are both $K$-by-$K$ matrices that only depend on angles and magnitudes, and hence rotational invariant. This leads to the following theorem characterizing the structure of out-of-plane critical points: 
\begin{theorem}
    If $d\ge K + 2$, then out-of-plane critical points (solutions of Eqn.~\ref{eq:normal-full}) are non-isolated and lie in a manifold. \label{thm:manifold}
\end{theorem}
The intuition is to construct a rotational matrix that is not identity matrix but keeps $\Pi_*$ invariant. Such matrices form a Lie group $\mathcal{L}$ that transforms critical points to critical points. Then for any out-of-plane critical point, there is one matrix in $\mathcal{L}$ that changes at least one of its weights, yielding a non-isolated different critical point. 

Note that Thm.~\ref{thm:manifold} also works for any general isotropic distribution, in which $\ee{F(\ve, \vw)}$ has the form of Eqn.~\ref{eq:f-generalized}. This is due to the symmetry of the input $X$, which in turn affects the geometry of critical points. The theorem also explains why we have flat minima~\newcite{hochreiter1995simplifying, dauphin2014identifying} often occuring in practice. 

\subsection{In-Plane Normal Equation}
To analyze in-plane critical points, it suffices to study gradient projections on $\Pi_*$. When $\{\vw_j\}$ is full-rank, the projections could be achieved by right-multiplying both sides by $\{\ve_{j'}\}$, which gives $K^2$ equations:
\begin{equation}
    M(\Theta) \normvw = M\opt(\Theta, \Theta\opt)\normvwopt \label{eq:normal} 
\end{equation}
This again shows decomposition of angles and magnitudes, and linearity with respect to the norms of weight vectors. Here $\normvw = [\|\vw_1\|, \|\vw_2\|, \ldots, \|\vw_K\|]\trans$ and similarly for $\normvwopt$. $M$ and $M\opt$ are $K^2$-by-$K$ matrices that only depend on angles. Entries of $M$ and $M\opt$ are:
\begin{eqnarray}
    m_{jj',k} &=&  (\pi - \theta^k_j) \cos\theta^k_{j'} + \sin\theta^k_j\cos\theta^j_{j'} \label{eq:m} \\
    m\opt_{jj',k} &=& (\pi - \theta^{*k}_j) \cos\theta^{*k}_{j'} + \sin\theta^{*k}_j\cos\theta^j_{j'} \label{eq:m-star} 
\end{eqnarray}
Here index $j$ is the $j$-th column of Eqn.~\ref{eq:normal-full}, $j'$ is from projection vector $\ve_{j'}$ and $k$ is the $k$-th weight magnitude.

\textbf{Diagnoal constraints}. For ``diagonal'' constraints $(j,j)$ of Eqn.~\ref{eq:normal}, we have $\cos\theta_j^j = 1$ and $m_{jj,k} = h(\theta^k_j)$, $m\opt_{jj,k} = h(\theta^{*k}_j)$, where $h(\theta) = (\pi - \theta) \cos\theta + \sin\theta$. Therefore, we arrive at the following subset of the constraints:
\begin{equation}
    M_r \normvw = M_r\opt\normvwopt \label{eq:linear-magnitude-reduced} 
\end{equation}
where $M_r = h(\Theta\trans)$ and $M\opt_r = h({\Theta\opt}\trans)$ are both $K$-by-$K$ matrices. Note that if $M_r$ is full-rank, then we could solve $\normvw$ from Eqn.~\ref{eq:linear-magnitude-reduced} and plug it back in Eqn.~\ref{eq:normal} to check whether it is indeed a critical point. This gives necessary conditions for critical points that only depend on angles.

\begin{figure}
    \includegraphics[width=0.5\textwidth]{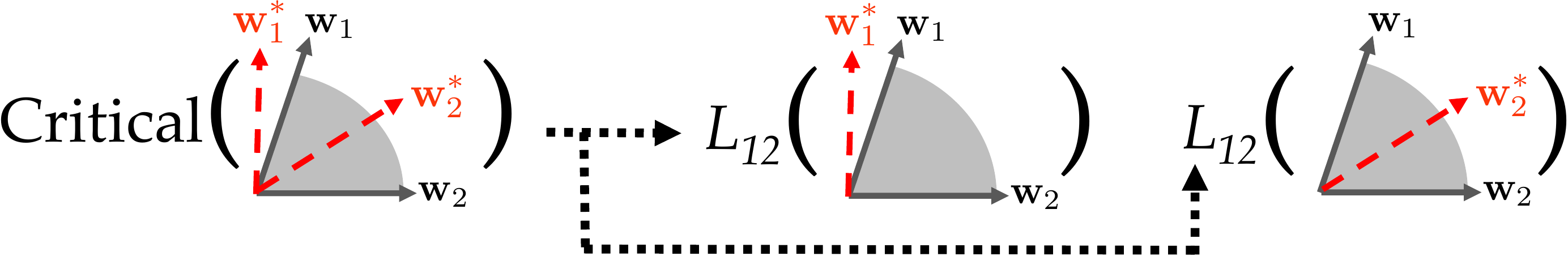}
    \vspace{-0.2in}
    \caption{Separable property of critical points using $L_{jj'}$ function (Eqn.~\ref{eq:l}). Checking the criticability of $\{\vw_1, \vw_2, \vw\opt_1, \vw\opt_2\}$ can be decomposed into two subproblems, one related to $\{\vw_1, \vw_2, \vw\opt_1\}$ and the other is related to $\{\vw_1, \vw_2, \vw\opt_2\}$.}\label{fig:separate-critical-points}
\end{figure}

\textbf{Separable Property}. Interestingly, the plugging back operation leads to conditions that are separable with respect to ground truth weight (Fig.~\ref{fig:separate-critical-points}). To see this, we first define the following quantity $L_{jj'}$ which is a function between \emph{a single} (rather than $K$) ground truth unit weight vector $\ve\opt$ and all current unit weights $\{\ve_l\}_{l=1}^K$: 
\begin{equation}
    L_{jj'}(\{\theta\opt_l\}, \Theta) = m\opt_{jj'} - \vv\trans M_r^{-1}\vm_{jj'} \label{eq:l} 
\end{equation}
where $\theta\opt_l = \angle(\ve\opt, \ve_l)$ is the angle between $\ve\opt$ and $\ve_l$, $\vv = \vv(\{\theta\opt_l\}) = [h(\theta\opt_1), \ldots, h(\theta\opt_K)]\trans$, and $m\opt_{jj'} = (\pi - \theta\opt_j) \cos\theta\opt_{j'} + \sin\theta\opt_j\cos\theta^j_{j'}$ (like Eqn.~\ref{eq:m-star}). Note that $\vv(\{\theta^{*j}_l\})$ is the $j$-th column of $M_r\opt$. Fig.~\ref{fig:separate-critical-points} illustrates the case when $K=2$. $L_{jj'}$ has the following properties:
\begin{proposition}
    $L_{jj'}(\{\theta\opt_l\}, \Theta) = 0$ when there exists $l$ so that $\ve\opt =\ve_l$. In addition, $L_{jj}(\{\theta\opt_l\}, \Theta) = 0$ always.
\end{proposition}
Intuitively, $L_{jj'}$ characterizes the relative geometric relationship among $\ve\opt$ and $\{\ve_l\}$. It is like determinant of a matrix whose columns are $\{\ve_l\}$ and $\ve\opt$. With $L_{jj'}$, we have the following necessary conditions for critical points: 
\begin{theorem}
    \label{thm:necessary-condition}
    If $\normvwopt \neq 0$, and for a given parameter $\vw$, $L_{jj'}(\{\theta^{*k}_l\}, \Theta) > 0$ (or $< 0$) for all $1 \le k \le K$, then $\vw$ cannot be a critical point.
\end{theorem}

\subsection{Case study: $K=2$ network} 
In this case, $M_r$ and $M\opt_r$ are $2$-by-$2$ matrices. Here we discuss the case that both $\vw_1$ and $\vw_2$ are in $\Pi_*$.

\textbf{Saddle points}. When $\theta^1_2 = 0$ ($\vw_1$ and $\vw_2$ are collinear), $M_r = \pi\vone\vone\trans$ is singular since $\ve_1$ and $\ve_2$ are identical. From Eqn.~\ref{eq:normal-full}, if $\theta^{*1}_1 = \theta^{*2}_1$, i.e., they are both aligned with the bisector angle of $\vw\opt_1$ and $\vw\opt_2$, and $\pi{\normvw}\trans\vone = h\left(\theta^{*1}_{*2}/2\right) (\normvwopt)\trans\vone$, then the current solution is a saddle point. Note that this gives one constraint for two weight magnitudes, and thus there exist infinite solutions. 

\begin{figure}
    \includegraphics[width=0.5\textwidth]{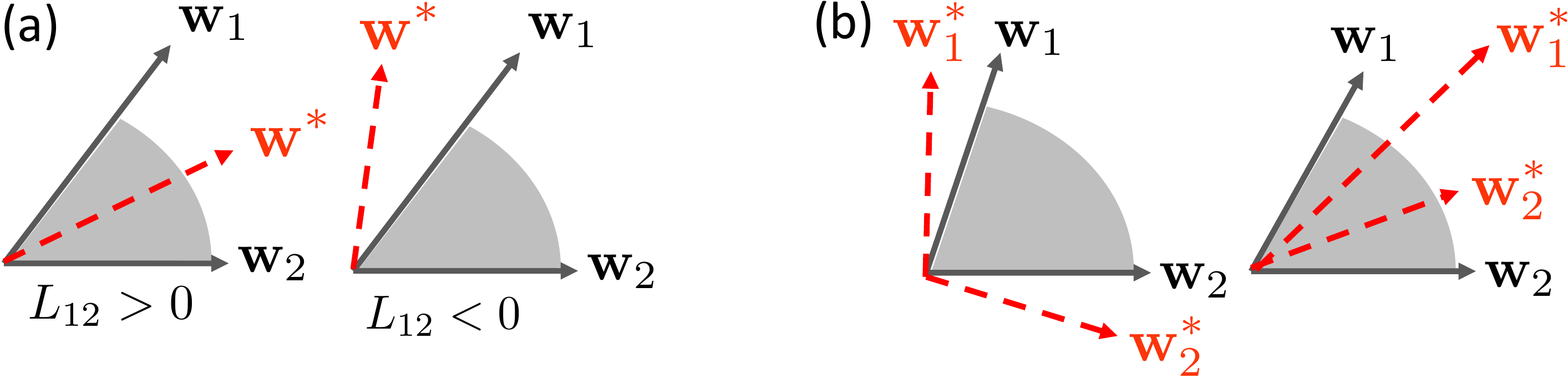}
    \vspace{-0.2in}
    \caption{Critical point analysis for $K=2$. \textbf{(a)} $L_{12}$ changes sign when $\vw\opt$ is in/out of the cone spanned by weights $\vw_1$ and $\vw_2$. \textbf{(b)} Two cases that $(\vw_1, \vw_2)$ cannot be critical points.}\label{fig:k2-critical-points}
\end{figure}

\textbf{Region without critical points}. We rely on the following conjecture that is verified empirically in an exhaustive manner (Sec.~\ref{sec:k2}). It characterizes zero-crossings of a 2D function on a closed region $[0, 2\pi]\times [0,\pi]$. In comparison, in-plane 2 ReLU network has $6$ parameters and is more difficult to handle: $8$ for $\vw_1$, $\vw_2$, $\vw\opt_1$ and $\vw\opt_2$, minus the rotational and scaling symmetries.
\begin{conjecture}
    \label{conj:2d}
    If $\ve\opt$ is in the interior of $\mathrm{Cone}(\ve_1, \ve_2)$, then $L_{12}(\theta\opt_1, \theta\opt_2, \theta^1_2) > 0$. If $\ve\opt$ is in the exterior, then $L_{12} < 0$.  
\end{conjecture}
This is also empirically true for $L_{21}$. Combined with Thm.~\ref{thm:necessary-condition}, we know that (Fig.~\ref{fig:k2-critical-points}):
\begin{theorem}
    If Conjecture~\ref{conj:2d} is correct, then for 2 ReLU network, $(\vw_1,\vw_2)$ ($\vw_1 \neq \vw_2$) is not a critical point, if they both are in $\mathrm{Cone}(\vw\opt_1, \vw\opt_2)$, or both out of it.
\end{theorem}
On the other hand, when exact one ground truth weight is inside $\mathrm{Cone}(\vw_1, \vw_2)$, it is not sure whether $(\vw_1, \vw_2)$ is a critical point.

\section{Convergence Analysis}
Application of Eqn.~\ref{eq:f} also yields interesting convergence analysis. We focus on infinitesimal analysis, i.e., when learning rate $\eta\rightarrow 0$ and the gradient update becomes a first-order differential equation:
\begin{equation}
    \dd\vw/\dd t = -\eex{\nabla_\vw J(\vw)} \label{eq:differential-equation}
\end{equation}
Then the populated objective $\eex{J}$ does not increase:
\begin{equation}
    \dd \ee{J} / \dd t = -\ee{\nabla J}\trans \dd\vw/\dd t = -\ee{\nabla J}\trans\ee{\nabla J} \le 0
\end{equation}
The goal of convergence analysis is to determine specific weight initializations $\vw\tone$ that leads to convergence to $\vw\opt$ following the gradient descent dynamics (Eqn.~\ref{eq:differential-equation}).

\subsection{Single ReLU case}
Using Lyapunov method~\newcite{lyapunov}, we show that the gradient dynamics (Eqn.~\ref{eq:differential-equation}) converges to $\vw\opt$ when $\vw\tone\in\Omega = \{\vw: \|\vw - \vw\opt\| < \|\vw\opt\|\}$:
\begin{theorem}
    When $\vw\tone\in\Omega = \{\vw : \|\vw - \vw\opt\| < \|\vw\opt\|\}$, following the dynamics of Eqn.~\ref{eq:differential-equation}, the Lyapunov function $V(\vw) = \frac{1}{2}\|\vw - \vw\opt\|^2$ has $\dd V/\dd t < 0$ and the system is asymptotically stable and thus $\vw\t\rightarrow\vw\opt$ when $t\rightarrow +\infty$.
\end{theorem}
The intuition is to represent $\dd V / \dd t$ as a $2$-by-$2$ bilinear form of vector $[\|\vw\|, \|\vw\opt\|]$, and the bilinear coefficient matrix, as a function of angles, is negative definite (except for $\vw = \vw\opt$). Note that similar approaches do not apply to regions including the origin because at the origin, the population gradient is discontinuous. $\Omega$ does not include the origin and for any initialization $\vw\tone \in \Omega$, we could always find a slightly smaller subset $\Omega'_\delta = \{\vw: \|\vw - \vw\opt\| \le \|\vw\opt\| - \delta\}$ with $\delta > 0$ that covers $\vw\tone$, and apply Lyapunov method within. Note that the global convergence claim in~\newcite{mei2016landscape} for $l_2$ loss does not apply to ReLU, since it requires $\sigma'(x) > 0$. 

\begin{figure}
    \includegraphics[width=0.5\textwidth]{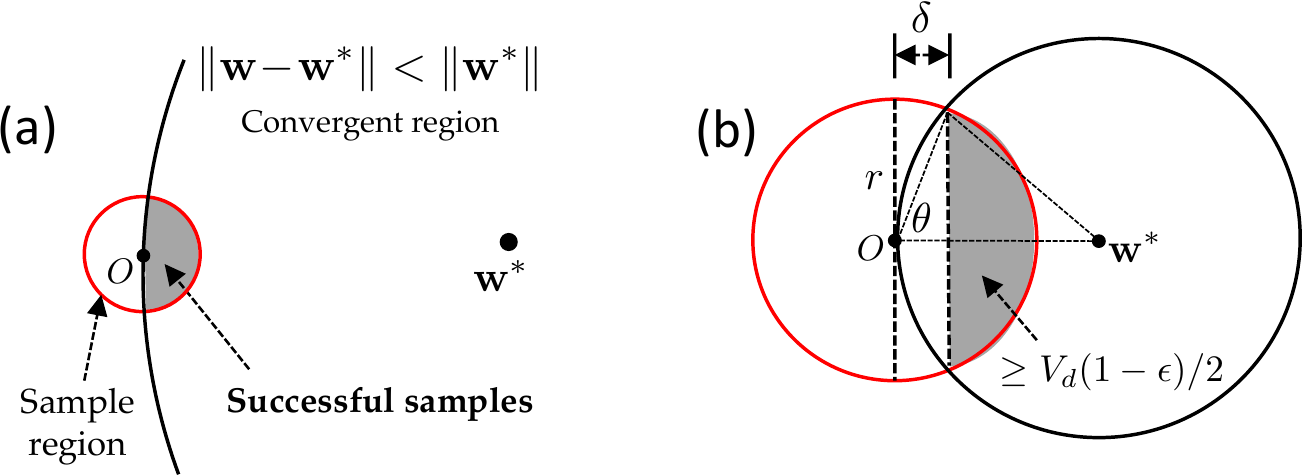}
    \vspace{-0.1in}
    \caption{\textbf{(a)} Sampling strategy to maximize the probability of convergence. \textbf{(b)} Relationship between sampling range $r$ and desired probability of success $(1 - \epsilon)/2$.} \label{fig:convergence-analysis}
\end{figure}

\textbf{Random Initialization.} How to sample $\vw\tone\in \Omega$ without knowing $\vw\opt$? Uniform sampling around origin with radius $r \ge \gamma\|\vw\opt\|$ for any $\gamma > 1$ results in exponentially small success rate $(r/\|\vw\opt\|)^d \le \gamma^{-d}$ in high-dimensional space. A better idea is to sample around the origin with very small radius (but not at $\vw = 0$), so that $\Omega$ looks like a hyperplane near the origin, and thus almost half samples are useful (Fig.~\ref{fig:convergence-analysis}(a)), as shown in the following theorem:
\begin{theorem}
    \label{thm:sampling}
    The dynamics in Eqn.~\ref{eq:dynamics} converges to $\vw\opt$ with probability at least $(1 - \epsilon)/2$, if the initial value $\vw\tone$ is sampled uniformly from $B_r = \{\vw : \|\vw\|\le r\}$ with $r \le \epsilon\sqrt{\frac{2\pi}{d + 1}}\|\vw\opt\|$. 
\end{theorem}
The idea is to lower-bound the probability of the shaded area (Fig.~\ref{fig:convergence-analysis}(b)). Thm.~\ref{thm:sampling} gives an explanation for common initialization techniques~\cite{xavier,PReLU, lecun2012efficient, bottou-88b} that uses random variables with $O(1/\sqrt{d})$ standard deviation. 

\subsection{Multiple ReLU case}
\label{sec:multiple-relus-case}
For multiple ReLUs, Lyapunov method on Eqn.~\ref{eq:two-layer-fix-top} yields no decisive conclusion. Here we focus on the symmetric property of Eqn.~\ref{eq:two-layer-fix-top} and discuss a special case, that the teacher parameters $\{\vw\opt_j\}_{j=1}^K$ and the initial weights $\{\vw\tone_j\}_{j=1}^K$ respect the following symmetry: $\vw_{j} = P_j\vw$ and $\vw\opt_{j} = P_j\vw\opt$, where $P_j$ is an orthogonal matrix whose collection $\mathcal{P} \equiv\{P_j\}_{j=1}^K$ forms a group. Without loss of generality, we set $P_1$ as the identity. Then from Eqn.~\ref{eq:two-layer-fix-top} the population gradient becomes: 
\begin{equation}
    \gradj = P_j \gradone
\end{equation}
This means that if all $\vw_j$ and $\vw\opt_j$ are symmetric under group actions, so does their population gradients. Therefore, the trajectory $\{\vw\t\}$ also respects the symmetry (i.e., $P_j\vw\t_1 = \vw\t_j$) and we only need to solve one equation for $\ee{\nabla_\vw J}$ instead of $K$ (here $\ve = \vw/\|\vw\|$):
\begin{equation}
    \ee{\nabla_\vw J} = \sum_{j'=1}^K \ee{F(\ve, P_{j'}\vw)} - \sum_{j'=1}^K \ee{F(\ve, P_{j'}\vw\opt)} \label{eq:symmetric-one} 
\end{equation}

\begin{figure}
    \includegraphics[width=0.5\textwidth]{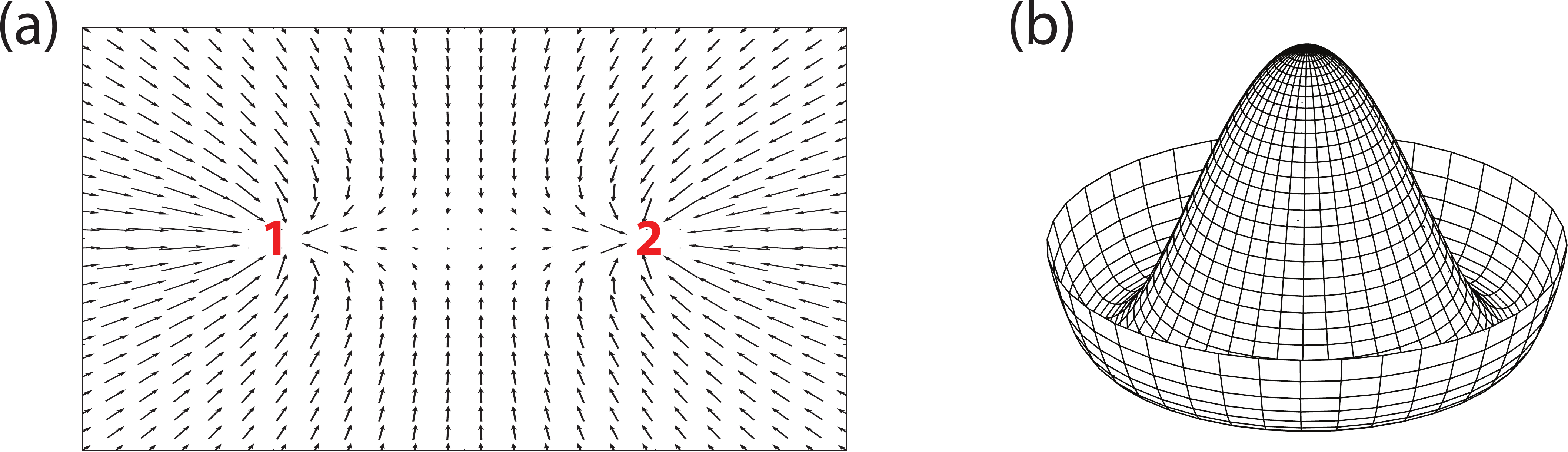}
    \vspace{-0.1in}
    \caption{Spontaneous Symmetric-Breaking (SSB): Objective / gradient field is symmetric but the solution is not. \textbf{(a)} Reflection keeps the gradient field invariant, but transforms $1$ to $2$ and vice versa. \textbf{(b)} The Mexican hat example. Rotation keeps the objective invariant, but transforms any local minimum to a different one.} \label{fig:ssb}
\end{figure}

Eqn.~\ref{eq:symmetric-one} has interesting properties, known as Spontaneous Symmetric-Breaking (SSB) in physics~\cite{brading2003symmetries}, in which the equations of motion respect a certain symmetry but its solution breaks it (Fig.~\ref{fig:ssb}). In our language, despite that the population gradient field $\ee{\nabla_\vw J}$ and the objective $\ee{J}$ are invariant to the group transformation $\mathcal{P}$, i.e., for $\vw\opt\rightarrow P_j\vw\opt$, $\ee{J}$ and $\ee{\nabla_\vw J}$ remain the same, its solution is not ($P_j\vw\neq \vw$). Furthermore, since $\mathcal{P}$ is finite, as we will see, the final solution converges to different permutations of $\vw\opt$ due to infinitesimal perturbations of initialization.

To illustrate such behaviors, consider the following example in which $\{\vw\opt_j\}_{j=1}^K$ forms an orthonormal basis and under this basis, $\mathcal{P}$ is a cyclic group in which $P_j$ circularly shifts dimension by $j - 1$ (e.g., $P_2 [1, 2, 3]\trans = [3, 1, 2]\trans$). In this case, if we start with $\vw\tone = x\tone\vw\opt + \sum_{j\neq 1}P_j\vw\opt_j = [x\tone, y\tone, \ldots, y\tone]$ under the basis of $\vw\opt$, then Eqn.~\ref{eq:symmetric-one} is further reduced to a convergent 2D nonlinear dynamics and Thm.~\ref{thm:symmetric-breaking} holds (Please check Supplementary Materials for the associated close-form of the 2D dynamics):
\begin{theorem}
\label{thm:symmetric-breaking}
For a bias-free two-layered ReLU network $g(\vx; \vw) = \sum_j \sigma(\vw_j\trans\vx)$ that takes spherical Gaussian inputs, if the teacher's parameters $\{\vw\opt_j\}$ form orthnomal bases, then (1) when the student parameters is initialized to be $[x\tone, y\tone, \ldots, y\tone]$ under the basis of $\vw\opt$, where $(x\tone, y\tone)\in \Omega = \{x\in(0,1], y\in[0,1], x>y\}$, then Eqn.~\ref{eq:two-layer-fix-top} converges to teacher's parameters $\{\vw\opt_j\}$ (or $(x,y)=(1,0)$); (2) when $x\tone = y\tone \in (0, 1]$, then it converges to a saddle point $x = y = \frac{1}{\pi K}(\sqrt{K-1} - \arccos(1/\sqrt{K}) + \pi)$. 
\end{theorem}
Thm.~\ref{thm:symmetric-breaking} suggests that when $\vw\tone = [y\tone, x\tone, \ldots, y\tone]$, the system converges to $P_2\vw\opt$, etc. Since $|x\tone - y\tone|$ can be arbitrarily small, a slightest perturbation around $x\tone=y\tone$ leads to a different fixed point $P_j\vw\opt$ for some $j$. Unlike single ReLU case, the initialization in Thm.~\ref{thm:symmetric-breaking} is $\vw\opt$-dependent, and serves as an example for the branching behavior.

Thm.~\ref{thm:symmetric-breaking} also suggests that for convergence, $x\tone$ and $y\tone$ can be arbitrarily small, regardless of the magnitude of $\vw\opt$, showing a global convergence behavior. In comparison, \newcite{saad1996dynamics} uses Gaussian error function ($\sigma = \mathrm{erf}$) as the activation, and only analyzes local behaviors near the two fixed points (origin and $\vw\opt$).

In practice, even with noisy initialization, Eqn.~\ref{eq:symmetric-one} and the original dynamics (Eqn.~\ref{eq:two-layer-fix-top}) still converge to $\vw\opt$ (and its transformations). We leave it as a conjecture, whose proof may lead to an initialization technique for 2-layered ReLU that is $\vw\opt$-independent.
\vspace{-0.1in}
\begin{conjecture}
    If the initialization $\vw\tone = x\tone \vw\opt + y\tone\sum_{j\neq 1}P_j \vw\opt + \boldsymbol{\epsilon}$, where $\boldsymbol{\epsilon}$ is noise and $(x\tone, y\tone) \in \Omega$, then Eqn.~\ref{eq:two-layer-fix-top} also converges to $\vw\opt$ with high probability.
\end{conjecture}

\section{Simulations}
\label{sec:simulation}
\begin{figure*}[t]
    \centering
    \includegraphics[width=0.95\textwidth]{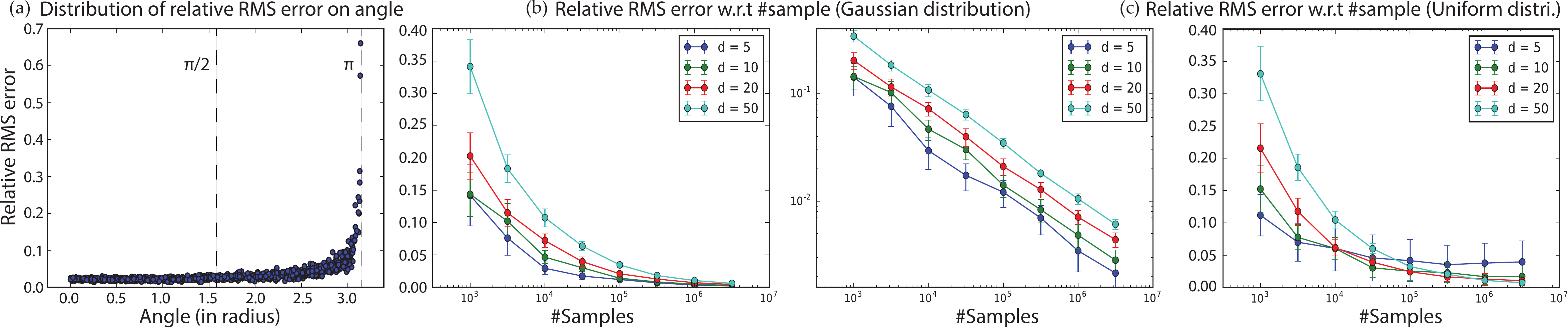}
    \vspace{-0.1in}
    \caption{\textbf{(a)} Distribution of relative RMS error with respect to $\theta = \angle(\vw, \ve)$. \textbf{(b)} Relative RMS error decreases with sample size, showing the asympototic behavior of the analytical formula (Eqn.~\ref{eq:f}). Note that the $y$-axis of the right plot is in log scale. \textbf{(c)} Eqn.~\ref{eq:f} also works well when the input data $X$ are generated by other zero-mean distribution $X$, e.g., uniform distribution in $[-1/2,1/2]$.} 
    \label{fig:one-layer-simulation}
\end{figure*}

\begin{figure*}[t]
    \centering
    \includegraphics[width=0.95\textwidth]{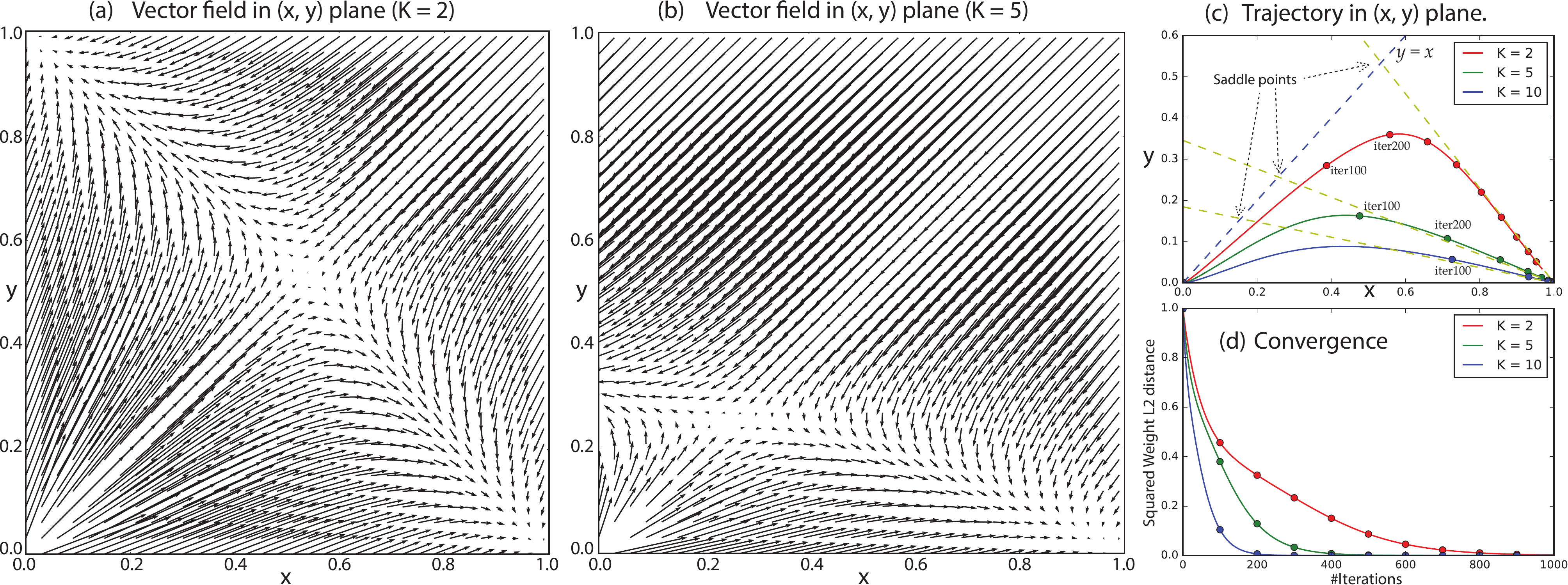}
    \vspace{-0.1in}
    \caption{\textbf{(a)-(b)} Vector field in $(x, y)$ plane following 2D dynamics (Thm.~\ref{thm:symmetric-breaking}, See Supplementary Materials for the close-form formula) for $K = 2$ and $K = 5$. Saddle points are visible. The parameters of teacher's network are at $\vw\opt=(1, 0)$. \textbf{(c)} Trajectory in $(x,y)$ plane for $K = 2$, $K = 5$, and $K=10$. All trajectories start from $\vw\tone=(10^{-3}, 0)$. Even $\vw\tone$ is aligned with $\vw\opt$, gradient descent takes detours. \textbf{(d)} Training curve. Interestingly, when $K$ is larger the convergence is faster.} 
    \label{fig:2d-dynamics}
\end{figure*}

\begin{figure*}[t]
    \centering
    \includegraphics[width=0.9\textwidth]{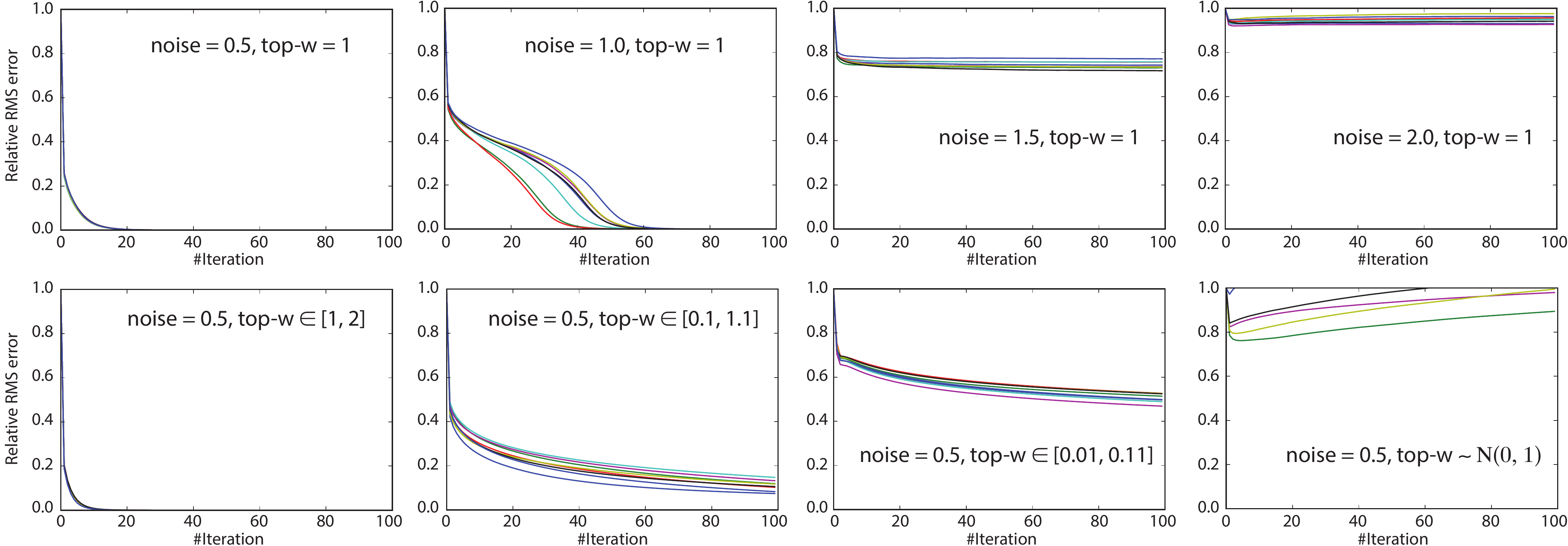}
    \vspace{-0.1in}
    \caption{\textbf{Top row:} Convergence when weights are initialized with noise: $\vw\tone = 10^{-3}\vw\opt + \epsilon$, where $\epsilon \sim N(0, 10^{-3}*noise)$. The 2-layered network converges to $\vw\opt$ until huge noise. Both teacher and student networks use $g(\vx) = \sum_{j=1}^K \sigma(\vw_j\trans\vx)$. Each experiment has 8 runs. \textbf{Bottom row:}  Convergence for $g_2(\vx) = \sum_{j=1}^K a_j\sigma(\vw_j\trans\vx)$. Here we fix top weights $a_j$ at different numbers (rather than $1$). Large positive $a_j$ corresponds to fast convergence. When $\{a_j\}$ contains mixture signs, convergence to $\vw\opt$ is not achieved.} \label{fig:empirical-two-layer-convergence}
\end{figure*}

\subsection{The analytical solution to $F(\ve, \vw)$}
We verify $\ee{F(\ve, \vw)} = \ee{X\trans D(\ve)D(\vw) X\vw}$ (Eqn.~\ref{eq:f}) with simulation. We randomly pick $\ve$ and $\vw$ so that their angle $\angle(\ve, \vw)$ is uniformly distributed in $[0, \pi]$. The analytical formula $\ee{F(\ve, \vw)}$ is compared with $F(\ve, \vw)$, which is computed via sampling on the input $X$ that follows spherical Gaussian distribution. We use relative RMS error: $err = \|\ee{F(\ve, \vw)} - F(\ve, \vw)\| / \|F(\ve, \vw)\|$. Fig.~\ref{fig:one-layer-simulation}(a) shows the error distribution with respect to angles. For small $\theta$, the gating function $D(\vw)$ and $D(\ve)$ mostly overlap and give a reliable estimation. When $\theta\rightarrow \pi$, $D(\vw)$ and $D(\ve)$overlap less and the variance grows. Note that our convergence analysis operate on $\theta\in [0, \pi/2]$ and is not affected. In the following, we sample angles from $[0, \pi/2]$.

Fig.~\ref{fig:one-layer-simulation}(a) shows that the accuracy of the formula increases with more samples. We also examine other zero-mean distributions of $X$, e.g., uniform distribution in $[-1/2, 1/2]$. Fig.~\ref{fig:one-layer-simulation}(d) shows that the formula still works for large $d$. Note that the error is computed up to a global scale, due to different normalization constants in probability distributions. To prove the usability of Eqn.~\ref{eq:f} for more general distributions remains open.

\begin{figure}
    \centering
    \includegraphics[width=0.47\textwidth]{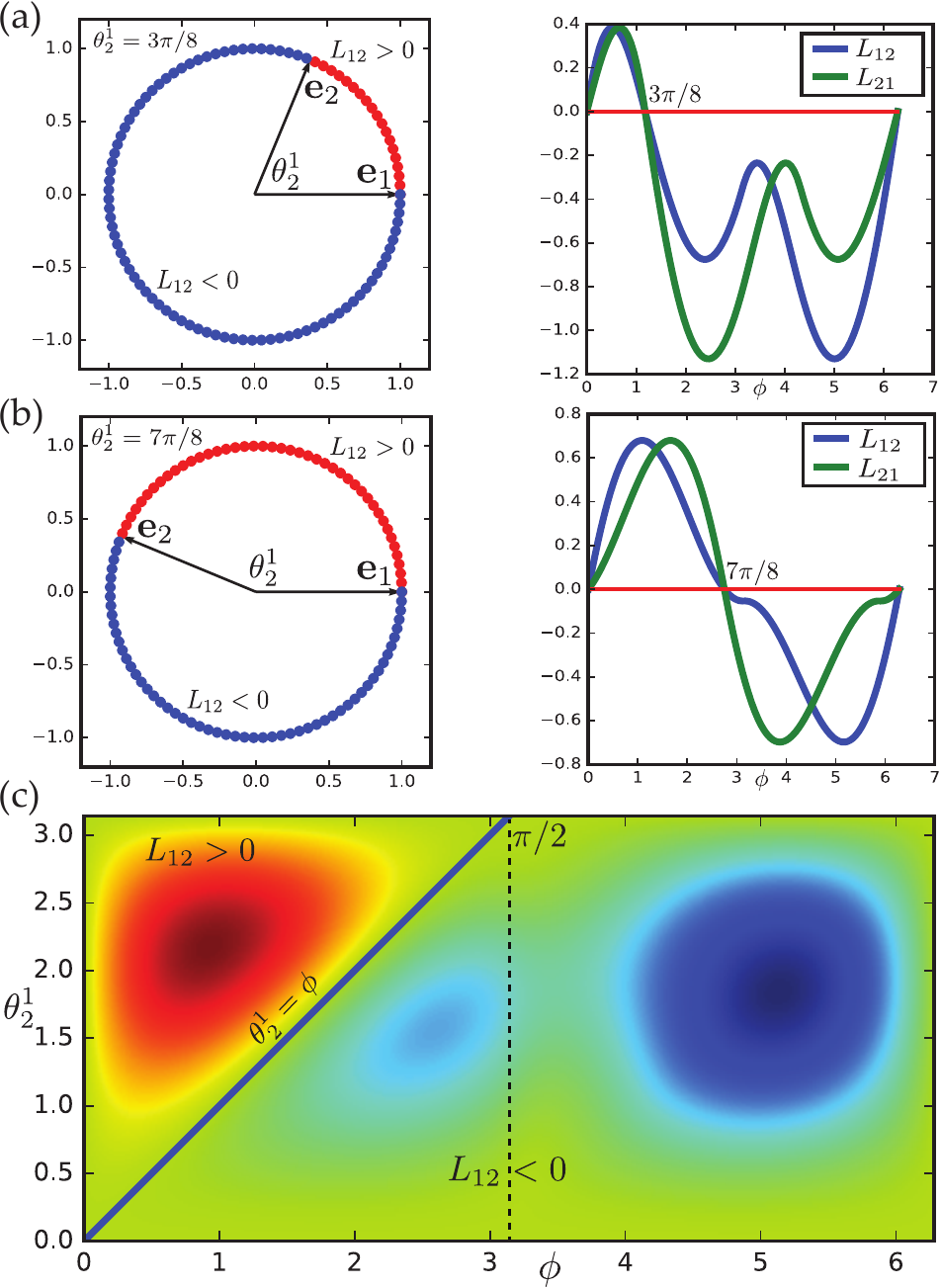}
    \caption{Quantity $L_{12}(\theta\opt_1, \theta\opt_2, \theta^1_2)$ and $L_{21}(\theta\opt_1, \theta\opt_2, \theta^1_2)$ in $2$ ReLU network. We fix $\theta^1_2 = \angle(\ve_1, \ve_2)$ and vary $\ve\opt = [\cos\phi, \sin\phi]\trans$. In this case, $\theta\opt_1$ and $\theta\opt_2$ are both dependent variables with respect to $\phi$. When $\ve\opt \in \mathrm{Cone}(\ve_1, \ve_2)$, $L_{12}$ and $L_{21} > 0$, otherwise negative. There are no extra zero-crossings. \textbf{(a)-(b)} Examples: $\theta^1_2 = 3\pi/8$ and $\theta^1_2 = 7\pi/8$. \textbf{(c)} Empirical evaluation on $(\theta^1_2, \phi) \in [0, \pi] \times [0, 2\pi]$ with grid size $10^4 \times 10^4$.} \label{fig:l12-empirical}
\end{figure}

\subsection{Empirical Results in critical point analysis $K=2$}
\label{sec:k2}
Conjecture~\ref{conj:2d} can be reduced to enumerate a complicated but 2D function via exhaustive sampling. In comparison, a full optimization of 2-ReLU network constrained on principal hyperplane $\Pi_*$ involves 6 parameters ($8$ parameters minus $2$ degrees of symmetry) and is more difficult to handle. Fig.~\ref{fig:l12-empirical} shows that empirically $L_{12}$ has no extra zero-crossing other than $\ve\opt = \ve_1$ or $\ve_2$. As shown in Fig.~\ref{fig:l12-empirical}(c), we have densely enumerated $\theta^1_2 \in [0, \pi]$ and $\ve\opt$ on a $10^4\times 10^4$ grid without finding any counterexamples.

\subsection{Convergence analysis for multiple ReLU nodes}
Fig.~\ref{fig:2d-dynamics}(a) and (b) shows the 2D vector field in Thm~\ref{thm:symmetric-breaking}. Fig.~\ref{fig:2d-dynamics}(c) shows the 2D trajectory towards convergence to the teacher's parameters $\vw\opt$. Interestingly, even when we initialize the weights as $[10^{-3}, 0]\trans$, whose direction is aligned with $\vw\opt$ at $[1, 0]\trans$, the gradient descent still takes detours to reach the destination. This is because at the beginning of optimization, all ReLU nodes explain the training error in the same way (both $x$ and $y$ increases); when the ``obvious'' component is explained, the error pushes some nodes to explain other components. Hence, specialization follows ($x$ increases but $y$ decreases). 

Fig.~\ref{fig:empirical-two-layer-convergence} shows empirical convergence for $K\ge 2$, when the initialization deviates from initialization $[x, y, \ldots, y]$ in Thm.~\ref{thm:symmetric-breaking}. Unless the deviation is large, $\vw$ converges to $\vw\opt$. For more general network $g_2(\vx) = \sum_{j=1}^K a_j\sigma(\vw_j\trans\vx)$, when $a_j > 0$ convergence follows. When some $a_j$ is negative, the network fails to converge to $\vw\opt$, even when the student is initialized with the true values $\{a\opt_j\}_{j=1}^K$. 

\section{Extension to multilayer ReLU network}
A natural question is whether the proposed method can be extended to multilayer ReLU network. In this case, there is similar subtraction structure for gradient as Eqn.~\ref{eq:basic-gradient}:
\begin{proposition}
\label{proposition:gradient-inflow}
Denote $[c]$ as all nodes in layer $c$. Denote $\vu\opt_j$ and $\vu_j$ as the output of node $j$ at layer $c$ of the teacher and student network, then the gradient of the parameters $\vw_j$ immediate under node $j \in [c]$ is:
\begin{equation}
    \nabla_{\vw_j} J = X_c\trans D_j Q_j \sum_{j'\in [c]} (Q_{j'}\vu_{j'} - Q\opt_{j'} \vu\opt_{j'})
\end{equation}
where $X_c$ is the data fed into node $j$, $Q_{j}$ and $Q\opt_{j}$ are $N$-by-$N$ diagonal matrices. For any node $k \in [c+1]$, $Q_k = \sum_{j \in [c]} w_{jk}D_jQ_j$ and similarly for $Q\opt_k$. 
\end{proposition}
The 2-layered network in this paper is a special case with $Q_j = Q\opt_j = I$. Despite the difficulty that $Q_j$ is now depends on the weights of upper layers, and the input $X_c$ is not necessarily Gaussian distributed, Proposition~\ref{proposition:gradient-inflow} gives a mathematical framework to explore the structure of gradient. For example, a similar definition of Population Gradient function is possible.

\section{Conclusion and Future Work}
In this paper, we study the gradient descent dynamics of a 2-layered bias-free ReLU network. The network is trained using gradient descent to reproduce the output of a teacher network with fixed parameters $\vw\opt$ in the sense of $l_2$ norm. We propose a novel analytic formula for population gradient when the input follows zero-mean spherical Gaussian distribution. This formula leads to interesting critical point and convergence analysis. Specifically, we show that critical points out of the hyperplane spanned by $\vw\opt$ are not isolated and form manifolds. For two ReLU case, we characterize regions that contain no critical points. For convergence analysis, we show guaranteed convergence for a single ReLU case with random initialization whose standard deviation is on the order of $O(1/\sqrt{d})$. For multiple ReLU case, we show that an infinitesimal change of weight initialization leads to convergence to different optima. 

Our work opens many future directions. First, Thm.~\ref{thm:manifold} characterizes the non-isolating nature of critical points in the case of isotropic input distribution, which explains why often practical solutions of NN are degenerated. What if the input distribution has different symmetries? Will such symmetries determine the geometry of critical points? Second, empirically we see convergence cases that are not covered by the theorems, suggesting the conditions imposed by the theorems can be weaker. Finally, how to apply similar analysis to broader distributions and how to generalize the analysis to multiple layers are also open problems. 

\textbf{Acknowledgement} We thank L{\'e}on Bottou, Ruoyu Sun, Jason Lee, Yann Dauphin and Nicolas Usunier for discussions and insightful suggestions.

\iffalse
To some extent, this setting covers the common one that a network is trained to fit a dataset, since we could replace the dataset witih a network with hidden parameters, by  

Note that this is a bit different from the common setting that a network is trained to fit a dataset. However, due to universal approximation theorem~\newcite{} of neural networks and its empirical flexibility, we could replace the dataset with a network with hidden parameters, train a student network and check whether the weight converges to the hidden parameters. 
\fi

\bibliography{references}

\begin{thebibliography}{24}
\providecommand{\natexlab}[1]{#1}
\providecommand{\url}[1]{\texttt{#1}}
\expandafter\ifx\csname urlstyle\endcsname\relax
  \providecommand{\doi}[1]{doi: #1}\else
  \providecommand{\doi}{doi: \begingroup \urlstyle{rm}\Url}\fi

\bibitem[Bottou(1988)]{bottou-88b}
Bottou, {L\'eon}.
\newblock Reconnaissance de la parole par reseaux connexionnistes.
\newblock In \emph{Proceedings of Neuro Nimes 88}, pp.\  197--218, Nimes,
  France, 1988.
\newblock URL \url{http://leon.bottou.org/papers/bottou-88b}.

\bibitem[Brading \& Castellani(2003)Brading and
  Castellani]{brading2003symmetries}
Brading, Katherine and Castellani, Elena.
\newblock \emph{Symmetries in physics: philosophical reflections}.
\newblock Cambridge University Press, 2003.

\bibitem[Choromanska et~al.(2015{\natexlab{a}})Choromanska, Henaff, Mathieu,
  Arous, and LeCun]{choromanska2015loss}
Choromanska, Anna, Henaff, Mikael, Mathieu, Michael, Arous, G{\'e}rard~Ben, and
  LeCun, Yann.
\newblock The loss surfaces of multilayer networks.
\newblock In \emph{AISTATS}, 2015{\natexlab{a}}.

\bibitem[Choromanska et~al.(2015{\natexlab{b}})Choromanska, LeCun, and
  Arous]{choromanska2015open}
Choromanska, Anna, LeCun, Yann, and Arous, G{\'e}rard~Ben.
\newblock Open problem: The landscape of the loss surfaces of multilayer
  networks.
\newblock In \emph{Proceedings of The 28th Conference on Learning Theory, COLT
  2015, Paris, France, July 3}, volume~6, pp.\  1756--1760, 2015{\natexlab{b}}.

\bibitem[Dauphin et~al.(2014)Dauphin, Pascanu, Gulcehre, Cho, Ganguli, and
  Bengio]{dauphin2014identifying}
Dauphin, Yann~N, Pascanu, Razvan, Gulcehre, Caglar, Cho, Kyunghyun, Ganguli,
  Surya, and Bengio, Yoshua.
\newblock Identifying and attacking the saddle point problem in
  high-dimensional non-convex optimization.
\newblock In \emph{Advances in neural information processing systems}, pp.\
  2933--2941, 2014.

\bibitem[Fukumizu \& Amari(2000)Fukumizu and Amari]{fukumizu2000local}
Fukumizu, Kenji and Amari, Shun-ichi.
\newblock Local minima and plateaus in hierarchical structures of multilayer
  perceptrons.
\newblock \emph{Neural Networks}, 13\penalty0 (3):\penalty0 317--327, 2000.

\bibitem[Glorot \& Bengio(2010)Glorot and Bengio]{xavier}
Glorot, Xavier and Bengio, Yoshua.
\newblock Understanding the difficulty of training deep feedforward neural
  networks.
\newblock In \emph{Aistats}, volume~9, pp.\  249--256, 2010.

\bibitem[He et~al.(2015)He, Zhang, Ren, and Sun]{PReLU}
He, Kaiming, Zhang, Xiangyu, Ren, Shaoqing, and Sun, Jian.
\newblock Delving deep into rectifiers: Surpassing human-level performance on
  imagenet classification.
\newblock In \emph{Proceedings of the IEEE International Conference on Computer
  Vision}, pp.\  1026--1034, 2015.

\bibitem[He et~al.(2016)He, Zhang, Ren, and Sun]{resnet}
He, Kaiming, Zhang, Xiangyu, Ren, Shaoqing, and Sun, Jian.
\newblock Deep residual learning for image recognition.
\newblock \emph{Computer Vision anad Pattern Recognition (CVPR)}, 2016.

\bibitem[Hinton et~al.(2012)Hinton, Deng, Yu, Dahl, Mohamed, Jaitly, Senior,
  Vanhoucke, Nguyen, Sainath, et~al.]{hinton2012deep}
Hinton, Geoffrey, Deng, Li, Yu, Dong, Dahl, George~E, Mohamed, Abdel-rahman,
  Jaitly, Navdeep, Senior, Andrew, Vanhoucke, Vincent, Nguyen, Patrick,
  Sainath, Tara~N, et~al.
\newblock Deep neural networks for acoustic modeling in speech recognition: The
  shared views of four research groups.
\newblock \emph{IEEE Signal Processing Magazine}, 29\penalty0 (6):\penalty0
  82--97, 2012.

\bibitem[Hochreiter et~al.(1995)Hochreiter, Schmidhuber,
  et~al.]{hochreiter1995simplifying}
Hochreiter, Sepp, Schmidhuber, J{\"u}rgen, et~al.
\newblock Simplifying neural nets by discovering flat minima.
\newblock \emph{Advances in Neural Information Processing Systems}, pp.\
  529--536, 1995.

\bibitem[Janzamin et~al.(2015)Janzamin, Sedghi, and
  Anandkumar]{janzamin2015beating}
Janzamin, Majid, Sedghi, Hanie, and Anandkumar, Anima.
\newblock Beating the perils of non-convexity: Guaranteed training of neural
  networks using tensor methods.
\newblock \emph{CoRR abs/1506.08473}, 2015.

\bibitem[Kawaguchi(2016)]{kawaguchi2016deep}
Kawaguchi, Kenji.
\newblock Deep learning without poor local minima.
\newblock \emph{Advances in Neural Information Processing Systems}, 2016.

\bibitem[Krizhevsky et~al.(2012)Krizhevsky, Sutskever, and Hinton]{alexnet}
Krizhevsky, Alex, Sutskever, Ilya, and Hinton, Geoffrey~E.
\newblock Imagenet classification with deep convolutional neural networks.
\newblock In \emph{Advances in neural information processing systems}, pp.\
  1097--1105, 2012.

\bibitem[LaSalle \& Lefschetz(1961)LaSalle and Lefschetz]{lyapunov}
LaSalle, J.~P. and Lefschetz, S.
\newblock Stability by lyapunov's second method with applications.
\newblock \emph{New York: Academic Press.}, 1961.

\bibitem[LeCun et~al.(2012)LeCun, Bottou, Orr, and
  M{\"u}ller]{lecun2012efficient}
LeCun, Yann~A, Bottou, L{\'e}on, Orr, Genevieve~B, and M{\"u}ller,
  Klaus-Robert.
\newblock Efficient backprop.
\newblock In \emph{Neural networks: Tricks of the trade}, pp.\  9--48.
  Springer, 2012.

\bibitem[Mei et~al.(2016)Mei, Bai, and Montanari]{mei2016landscape}
Mei, Song, Bai, Yu, and Montanari, Andrea.
\newblock The landscape of empirical risk for non-convex losses.
\newblock \emph{arXiv preprint arXiv:1607.06534}, 2016.

\bibitem[Saad \& Solla(1996)Saad and Solla]{saad1996dynamics}
Saad, David and Solla, Sara~A.
\newblock Dynamics of on-line gradient descent learning for multilayer neural
  networks.
\newblock \emph{Advances in Neural Information Processing Systems}, pp.\
  302--308, 1996.

\bibitem[Saxe et~al.(2013)Saxe, McClelland, and Ganguli]{saxe2013exact}
Saxe, Andrew~M, McClelland, James~L, and Ganguli, Surya.
\newblock Exact solutions to the nonlinear dynamics of learning in deep linear
  neural networks.
\newblock \emph{arXiv preprint arXiv:1312.6120}, 2013.

\bibitem[Simonyan \& Zisserman(2015)Simonyan and Zisserman]{vgg}
Simonyan, Karen and Zisserman, Andrew.
\newblock Very deep convolutional networks for large-scale image recognition.
\newblock \emph{International Conference on Learning Representations (ICLR)},
  2015.

\bibitem[Soudry \& Carmon(2016)Soudry and Carmon]{soudry2016no}
Soudry, Daniel and Carmon, Yair.
\newblock No bad local minima: Data independent training error guarantees for
  multilayer neural networks.
\newblock \emph{arXiv preprint arXiv:1605.08361}, 2016.

\bibitem[Sutskever et~al.(2014)Sutskever, Vinyals, and
  Le]{sutskever2014sequence}
Sutskever, Ilya, Vinyals, Oriol, and Le, Quoc~V.
\newblock Sequence to sequence learning with neural networks.
\newblock In \emph{Advances in neural information processing systems}, pp.\
  3104--3112, 2014.

\bibitem[Szegedy et~al.(2015)Szegedy, Liu, Jia, Sermanet, Reed, Anguelov,
  Erhan, Vanhoucke, and Rabinovich]{inception}
Szegedy, Christian, Liu, Wei, Jia, Yangqing, Sermanet, Pierre, Reed, Scott,
  Anguelov, Dragomir, Erhan, Dumitru, Vanhoucke, Vincent, and Rabinovich,
  Andrew.
\newblock Going deeper with convolutions.
\newblock In \emph{Computer Vision and Pattern Recognition (CVPR)}, pp.\  1--9,
  2015.

\bibitem[Zhang et~al.(2017)Zhang, Panigrahy, and Sachdeva]{eletron-proton}
Zhang, Qiuyi, Panigrahy, Rina, and Sachdeva, Sushant.
\newblock Electron-proton dynamics in deep learning.
\newblock \emph{arXiv preprint arXiv:1702.00458}, 2017.

\end{thebibliography}
\bibliographystyle{icml2017}

\end{document}

\maketitle

\section{Introduction}
No theorem is provided.

\section{Related Works}
No theorem is provided.

\section{Problem Definition}
No theorem is provided.

\section{The Analytical Formula}
Here we list all detailed proof for all the theorems. 
\subsection{One ReLU Case}
\begin{figure}
    \centering
    \includegraphics[width=.7\textwidth]{analytic2-crop.pdf}
    \caption{\textbf{(a)-(b)} Two cases in Thm.~\ref{thm:f-supp}. }\label{fig:f}
\end{figure}

\begin{theorem}
    \label{thm:f-supp}
    Suppose $F(\ve, \vw) = X\trans D(\ve) D(\vw) X\vw$ where $\ve$ is a unit vector and $X = [\vx_1, \vx_2, \cdots, \vx_N]\trans$ is $N$-by-$d$ sample matrix. If $\vx_i \sim N(0, I)$, then:
    \begin{equation}
        \ee{F(\ve, \vw)} = \frac{N}{2\pi} \left( (\pi - \theta)\vw + \lenw\sin\theta\ve\right) \label{eq:f-supp}
    \end{equation}
    where $\theta \in [0, \pi]$ is the angle between $\ve$ and $\vw$. 
\end{theorem}
\begin{proof}
    Note that $F$ can be written in the following form:
    \begin{equation}
        F(\ve, \vw) = \sum_{i: \vx_i\trans\ve \ge 0, \vx_i\trans\vw \ge 0} \vx_i \vx_i\trans\vw
    \end{equation}
    where $\vx_i$ are samples so that $X = [\vx_1, \vx_2, \cdots, \vx_n]\trans$. We set up the axes related to $\ve$ and $\vw$ as in Fig.~\ref{fig:f}, while the rest of the axis are prependicular to the plane. In this coordinate system, any vector $\vx = [r\sin\phi, r\cos\phi, x_3, \ldots, x_d]$. We have an orthonomal set of bases: $\ve$, $\ve_\perp = -\frac{\ve- \vw/\lenw\cos\theta}{\sin\theta}$ (and any set of bases that span the rest of the space). Under the basis, the representation for $\ve$ and $\vw$ is $[1,\vzero_{d-1}]$ and $[\lenw\cos\theta, -\lenw\sin\theta, \vzero_{d-2}]$. Note that here $\theta \in (-\pi, \pi]$. The angle $\theta$ is positive when $\ve$ ``chases after'' $\vw$, and is otherwise negative.

Now we consider the quality $R(\phi_0) = \ee{\frac{1}{N}\sum_{i: \phi_i \in [0, \phi_0]}\vx_i \vx_i\trans}$. If we take the expectation and use polar coordinate only in the first two dimensions, we have:
\begin{eqnarray}
    & & R(\phi_0) = \ee{\frac{1}{N}\sum_{i: \phi_i \in [0, \phi_0]}\vx_i \vx_i\trans} = \ee{\vx_i\vx_i\trans | \phi_i \in [0, \phi_0]} \pr{\phi_i \in [0, \phi_0]} \nonumber \\
    &=&\int_{0}^{+\infty} \iint_{-\infty}^{+\infty} \int_{0}^{\phi_0} 
    \begin{bmatrix}
        r\sin\phi \\
        r\cos\phi \\
        \ldots\\
        x_d
    \end{bmatrix}
    \begin{bmatrix}
        r\sin\phi & r\cos\phi & \ldots & x_d
    \end{bmatrix}
    p(r)p(\theta)\prod_{k=3}^d p(x_k)
    r\dd r\dd\phi \dd x_3\ldots \dd x_d\nonumber
\end{eqnarray}
where $p(r) = e^{-r^2/2}$ and $p(\theta) = 1 / 2\pi$. Note that $R(\phi_0)$ is a $d$-by-$d$ matrix. The first $2$-by-$2$ block can be computed in close form (note that $\int_{0}^{+\infty} r^2 p(r) r\dd r = 2$). Any off-diagonal element except for the first $2$-by-$2$ block is zero due to symmetric property of spherical Gaussian variables. Any diagonal element outside the first $2$-by-$2$ block will be $\pr{\phi_i \in [0, \phi_0]} = \phi_0 / 2\pi$. Finally, we have: 
\begin{eqnarray}
    R(\phi_0) &=& \ee{\frac{1}{N}\sum_{i: \phi_i \in [0, \phi_0]}\vx_i \vx_i\trans} = \frac{1}{4\pi}
    \begin{bmatrix}
        2\phi_0 - \sin2\phi_0 & 1 - \cos2\phi_0 & \vzero \\
        1 - \cos2\phi_0 & 2\phi_0 + \sin2\phi_0 & \vzero \\
        \vzero  & \vzero & 2\phi_0 I_{d-2}
    \end{bmatrix} \\ 
    &=& \frac{\phi_0}{2\pi}I_d + \frac{1}{4\pi}
    \begin{bmatrix}
        -\sin2\phi_0 & 1 - \cos2\phi_0 & \vzero \\
        1 - \cos2\phi_0 & \sin2\phi_0 & \vzero \\
        \vzero  & \vzero & 0
    \end{bmatrix}
\end{eqnarray}

With this equation, we could then compute $\ee{F(\ve, \vw)}$. When $\theta \ge 0$, the condition $\{i: \vx_i\trans\ve \ge 0, \vx_i\trans\vw \ge 0\}$ is equivalent to $\{i: \phi_i \in [\theta, \pi]\}$ (Fig.~\ref{fig:f}(a)). Using $\vw = [\lenw\cos\theta, -\lenw\sin\theta, \vzero_{d-2}]$ and we have: 
 \begin{eqnarray}
     & & \ee{F(\ve, \vw)} = N\left(R(\pi) - R(\theta)\right)\vw\\
     &=& \frac{N}{4\pi}\left(2(\pi - \theta)\vw - \lenw
    \begin{bmatrix}
        - \sin2\theta & 1 - \cos2\theta & \vzero \\
        1 - \cos2\theta & \sin2\theta & \vzero \\
        \vzero  & \vzero & 0
    \end{bmatrix}
    \begin{bmatrix}
        \cos\theta \\
        -\sin\theta \\
        \vzero
    \end{bmatrix}\right) \\
    &=& \frac{N}{2\pi} \left( (\pi-\theta)\vw + \lenw 
    \begin{bmatrix}
        \sin\theta \\
        \vzero
    \end{bmatrix}
    \right) \\
    &=& \frac{N}{2\pi} \left( (\pi-\theta)\vw + \lenw\sin\theta\ve\right)  
\end{eqnarray}
For $\theta < 0$, the condition $\{i: \vx_i\trans\ve \ge 0, \vx_i\trans\vw \ge 0\}$ is equivalent to $\{i: \phi_i \in [0, \pi+\theta]\}$ (Fig.~\ref{fig:f}(b)), and similarly we get
\begin{equation}
    \ee{F(\ve, \vw)} = N\left(R(\pi+\theta) - R(0)\right)\vw = \frac{N}{2\pi} \left( (\pi+\theta)\vw - \lenw\sin\theta\ve\right)  
\end{equation}
Notice that by abuse of notation, the $\theta$ appears in Eqn.~\ref{eq:f-supp} is the absolute value and Eqn.~\ref{eq:f-supp} follows.
\qed
\end{proof}

\section{Critical Point Analysis}
Remember that we have: suppose $F(\ve, \vw) = X\trans D(\ve) D(\vw) X\vw$ where $\ve$ is a unit vector and $X = [\vx_1, \vx_2, \cdots, \vx_N]\trans$ is $N$-by-$d$ sample matrix. If $\vx_i \sim N(0, I)$, then:
\begin{equation}
        \ee{F(\ve, \vw)} = \frac{N}{2\pi} \left( (\pi - \theta)\vw + \lenw\sin\theta\ve\right) \label{eq:f}
\end{equation}
where $\theta \in [0, \pi]$ is the angle between $\ve$ and $\vw$. And the expected gradient for $g(\vx) = \sum_{j=1}^K \sigma(\vw\trans_j\vx)$ with respect to $\vw_j$ is the following:
\begin{equation}
    - \ee{\nabla_{\vw_j} J} = \sum_{j'=1}^K \ee{F(\ve_j, \vw\opt_{j'})} - \ee{F(\ve_j, \vw_{j'})} \label{eq:two-layer-fix-top} 
\end{equation}
where $\ve_j = \vw_j / \|\vw_j\|$. By solving Eqn.~\ref{eq:two-layer-fix-top} ($\ee{\nabla_{\vw_j} J} = 0\quad j=1,\ldots,K$), it is possible to identify all critical points of $g(\vx)$. In general Eqn.~\ref{eq:two-layer-fix-top} is highly nonlinear and cannot be solved efficiently, however, we show that in our particular case, it is possible since Eqn.~\ref{eq:two-layer-fix-top} has interesting properties.

\def\normvw{\bar\vw}
\def\normvwopt{\bar\vw\opt}

First of all, the system of equations 
\begin{equation}
    \ee{\nabla_{\vw_j} J} = 0\quad, j=1,\ldots,K
\end{equation}
or 
\begin{equation}
    \sum_{j'=1}^K\ee{F(\ve_j, \vw_{j'}} = \sum_{j'=1}^K\ee{F(\ve_j, \vw\opt_{j'}}\quad, j=1,\ldots,K
\end{equation}
is a linear combination of $\vw_j$ and $\vw\opt_j$ but with varying coefficients. We could rewrite the system as follows:
\begin{eqnarray}
    \diag\va E\trans + B\diag\normvw E\trans = \diag{\va\opt} E\trans + B\opt {W\opt}\trans \label{eq:normal}
\end{eqnarray}
where $E$, $W$, $W\opt$, $\va$, $B$, $\va\opt$ and $B\opt$ are all $K$-by-$K$ matrices:
\begin{eqnarray}
    \va = \sin\Theta\trans \normvw &,\quad\quad & \va\opt = \sin{\Theta\opt}\trans \normvwopt \\
    B = \pi\vone\vone\trans - \Theta\trans &,\quad\quad & B\opt = \pi\vone\vone\trans - {\Theta\opt}\trans \\
    E = \left[\ve_1, \ve_2, \ldots, \ve_K\right] & & \\
    W = \left[\vw_1, \vw_2, \ldots, \vw_K\right] &,\quad\quad & W\opt = \left[\vw\opt_1, \vw\opt_2, \ldots, \vw\opt_K\right]
\end{eqnarray}
where $\theta_j^{*j'} \equiv \angle (\vw_j, \vw\opt_{j'})$ and $\theta_{j'}^j = \theta_j^{j'} \equiv \angle (\vw_j,\vw_{j'})$, $\Theta = [\theta^i_j]$ (the element at $i$-th row, $j$-th column of $\Theta$ is $\theta^i_j$) and $\Theta\opt = [\theta^{*i}_j]$, $\normvw = \left[\|\vw_1\|, \|\vw_2\|, \ldots, \|\vw_K\|\right]$ and $\normvwopt = \left[\|\vw\opt_1\|, \|\vw\opt_2\|, \ldots, \|\vw\opt_K\|\right]$.

Eqn.~\ref{eq:normal} already has interesting properties. The first thing we consider is whether the critical point will fall outside the \emph{Principle Hyperplane} $\Pi_*$, which is the plane spanned by the ground truth weight vectors $W\opt$. The following theorem shows that the critical points outside $\Pi_*$ must lie in a manifold:

\begin{lemma}
    If $\{\vw_j\}$ is a critical point satisfying Eqn.~\ref{eq:normal}, then for any orthogonal mapping $R$ with $R|_{\Pi_*} = Id$, $\{R\vw_j\}$ is also a critical point.
\end{lemma}
\begin{proof}
    First of all, since $R$ is an orthogonal transformation, it keeps all angles and magnitudes and $\va$, $\va\opt$, $B$, $\normvw$ and $\normvwopt$ are invariant. For simplicity we write $Y = \diag\va + B\diag\normvw - \diag{\va\opt}$ and $Y$ is also invariant under $R$. Since $R|_{\Pi_*} = Id$, we have $R W\opt = W\opt$ and
    \begin{equation}
        Y_R E_R\trans - B_R\opt{W\opt}\trans = Y E\trans R\trans - B\opt{W\opt}\trans R\trans = \left(Y E\trans- B\opt{W\opt}\trans\right)R\trans = 0
    \end{equation}
    \qed
\end{proof}
Note that for $d \ge K + 2$, there always exists $R \neq Id$ and satisfy such a condition, which yield continuous critical points. Further, such a transformation forms a Lie group. Therefore we have:
\begin{theorem}
    If $d\ge K + 2$, then any critical point satisfying Eqn.~\ref{eq:normal} and is outside $\Pi_*$ must lie in a manifold. 
\end{theorem}
The intuition is simple. For any out-of-plane critical point, pick a matrix that satisfies the condition of the theorem, and tranforms it to a different yet infinitely close critical points. Such a matrix always exists, since for the $d-K$ subspace, if it is odd, then we can always pick a rotation whose fixed axis is not aligned with all $K$ weights; if it is even, then there is a rotation matrix without a fixed point.

\subsection{Characteristics within the Principle Plane}
We could right-multiple $E$ and turn the normal equation to a linear function with respect to the magnitude of weights $\|\vw\|$. Note that we have:
\begin{equation}
    E\trans E = \cos\Theta,\quad\quad W\trans E = \diag\normvw\cos\Theta,\quad\quad (W\opt)\trans E = \diag\normvwopt\Theta\opt
\end{equation}

Therefore, Eqn.~\ref{eq:normal} becomes:
\begin{equation}
    \diag{\va}\cos\Theta + B\diag\normvw\cos\Theta = \diag{\va\opt}\cos\Theta + B\opt\diag\normvwopt\cos\Theta\opt
\end{equation}
which is a homogenous linear equation with respect to the magnitude of the weights (note that $\va$ and $\va\opt$ is linear to the magnitudes). In particular, the $(i,j)$ entry of the LHS and RHS of this equality are:
\begin{eqnarray}
    LHS_{ij} &=& \cos\theta^i_j \left(\sum_{k=1}^K \sin\theta_i^{k}\|\vw_{k}\|\right) + \sum_{k=1}^K (\pi - \theta^{k}_i)\|\vw_{k}\|\cos\theta^{k}_j \\
    RHS_{ij} &=& \cos\theta^i_j \left(\sum_{k=1}^K \sin\theta_i^{*k}\|\vw\opt_{k}\|\right) + \sum_{k=1}^K (\pi - \theta^{*k}_i)\|\vw\opt_{k}\|\cos\theta^{*k}_j 
\end{eqnarray}

Therefore, the following equation holds:
\begin{equation}
    M \normvw = M\opt\normvwopt \label{eq:linear-magnitude} 
\end{equation}
where $M$ and $M\opt$ are $K^2$-by-$K$ matrices. Each entry $m_{ij,k}$ that correponds to the coefficient of $k$-th weight vector at $(i,j)$ entry of Eqn.~\ref{eq:normal} is defined as:
\begin{eqnarray}
    m_{ij,k} &=&  (\pi - \theta^k_i) \cos\theta^k_j + \sin\theta^k_i\cos\theta^i_j \label{eq:m} \\
    m\opt_{ij,k} &=& (\pi - \theta^{*k}_i) \cos\theta^{*k}_j + \sin\theta^{*k}_i\cos\theta^i_j \label{eq:m-star} 
\end{eqnarray}
\textbf{Special case on the diagonal}. For diagonal element $(i,i)$, $\cos\theta_i^i = 1$ and $m_{ii,k} = h(\theta^k_i)$, $m\opt_{ii,k} = h(\theta^{*k}_i)$, where 
\begin{equation}
    h(\theta) = (\pi - \theta) \cos\theta + \sin\theta .
\end{equation} 
Therefore, with only diagonal element, we arrive at the following subset of the constraints to be satisfied for any critical points:
\begin{equation}
    M_r \normvw = M_r\opt\normvwopt \label{eq:linear-magnitude-reduced} 
\end{equation}
where $M_r = h(\Theta\trans)$ and $M\opt_r = h({\Theta\opt}\trans)$ are both $K$-by-$K$ matrices. Note that if $M_r$ is full-rank, then we could solve $\normvw$ from Eqn.~\ref{eq:linear-magnitude-reduced} and plug it back in Eqn.~\ref{eq:linear-magnitude} to check whether it is indeed a critical point. 

\begin{lemma}
    If $\normvwopt \neq 0$ (no trivial ground truth solutions), and for a given $(\Theta, \Theta\opt)$, there exists a row (e.g. $l$-th row) of $M$ and $M\opt$, namely $\vm\trans_l$ and ${\vm_l\opt}\trans$, satisfying
\begin{equation}
    \vm\opt_l - {M\opt_r}\trans M_r^{-1}\vm_l > 0 \quad \mathrm{or} \quad \vm\opt_l - {M\opt_r}\trans M_r^{-1}\vm_l < 0 \label{eq:one-line-condition} 
\end{equation}
Then $(\Theta, \Theta\opt)$ cannot be a critical point.
\end{lemma}
\begin{proof}
    Suppose given $(\Theta, \Theta\opt)$, we get $M_r$ and $M\opt_r$ and compute $\normvw$ using Eqn.~\ref{eq:linear-magnitude-reduced}, then we have 
    \begin{equation}
        (\normvwopt)\trans {M\opt_r}\trans M_r^{-1}\vm_l = (M\opt_r\normvwopt)\trans M_r^{-1}\vm_l = \normvw\trans M_r\trans M_r^{-1}\vm_l = \normvw\trans\vm_l
    \end{equation}
    Therefore, from the condition $\vm\opt_l - {M\opt_r}\trans M_r^{-1}\vm_l > 0$ and $\normvwopt \ge 0$ but $\normvwopt \neq 0$, we have
    \begin{equation}
        (\normvwopt)\trans(\vm\opt_l - {M\opt_r}\trans M_r^{-1}\vm_l) = (\normvwopt)\trans\vm\opt_l - \normvw\trans\vm_l > 0
    \end{equation}
    but this contradicts with the necessary condition for $(\Theta, \Theta\opt)$ to become a critical point (Eqn.~\ref{eq:linear-magnitude}). Similarly we can prove the other side. \qed
\end{proof}

\textbf{Separation property of Eqn.~\ref{eq:one-line-condition}}. Note that both the $k$-th element of $\vm\opt_l$ and ${M\opt_r}\trans M_r^{-1}\vm_l$ in Eqn.~\ref{eq:one-line-condition} are only dependent on the $k$-th true weight vector $\vw\opt_k$ (and all $\{\vw_j\}$). 
\begin{itemize}
    \item For $\vm\opt_l$, this can be seen by Eqn.~\ref{eq:m-star}, in which the $k$-th element is only related to the angles $\theta^{*k}_\cdot$ between $\vw\opt_k$ and $\{\vw_j\}$. 
    \item For ${M\opt_r}\trans M_r^{-1}\vm_l$, notice that the $k$-th column of $M\opt_r$ (the $k$-th row of ${M\opt_r}\trans$) is only related to $\vw\opt_k$ but not other ground truth weight vectors. This separation property makes analysis much easier, as shown in the case of $K=2$.
\end{itemize}
Therefore, we could consider the following function regarding to one (rather than $K$) ground truth unit weight vector $\ve\opt$ and all estimated unit vectors $\{\ve_l\}$:
\begin{equation}
    L_{ij}(\ve\opt, \{\ve_l\}) = m\opt_{ij} - {\vv\opt}\trans M_r^{-1}\vm_{ij} 
\end{equation}
where ${\vv\opt}\trans = [h(\theta\opt_1), h(\theta\opt_2), \ldots, h(\theta\opt_K)]$, $\theta\opt_j = \angle(\ve\opt, \vw_j)$ and $m\opt_{ij} = (\pi - \theta\opt_i) \cos\theta\opt_j + \sin\theta\opt_i\cos\theta^i_j$ (like Eqn.~\ref{eq:m-star}).

\begin{proposition}
    \label{proposition:property-L}
    $L_{ij}(\ve\opt, \{\ve_l\}) = 0$ for any $\ve\opt = \ve_l$, $1\le l\le K$. In addition, $L_{ii}(\ve\opt, \{\ve_l\}) \equiv 0$.
\end{proposition}
\begin{proof}
    When $\ve\opt = \ve_l$, then $\theta\opt_k = \theta^l_k$ and ${\vv\opt}\trans$ becomes the $l$-th row of $M_r$. Since $M_rM_r^{-1} = I_{K\times K}$, ${\vv\opt}\trans M_r^{-1}$ becomes a unit vector with only $l$-th element being $1$. Therefore, again with $\theta\opt_k = \theta^l_k$, we have:
    \begin{equation}
        L_{ij}(\ve\opt, \{\ve_l\}) = m\opt_{ij} - m_{ij,l} = 0 
    \end{equation}
    For $L_{ii}$, by definition $\vm_{ii}$ is $i$-th column of $M_r$, so $M_r^{-1}\vm_{ii}$ is a unit vector with only $i$-th element being 1. Therefore
    \begin{equation}
        (\vv\opt)\trans M_r^{-1}\vm_{ii} = h(\theta\opt_i) = m\opt_{ii}
    \end{equation}
    \qed 
\end{proof}

Then the previous lemma can be written as the following:
\begin{theorem}
    \label{thm:necessary-condition}
    If $\normvwopt \neq 0$, and for a given parameter $\vw$, $L_{jj'}(\{\theta^{*k}_l\}, \Theta) > 0$ or $< 0$ for all $1 \le k \le K$, then $\vw$ cannot be a critical point.
\end{theorem}

\subsection{ReLU network with two hidden nodes ($K=2$)}
For $K = 2$, we have $4$-by-$2$ matrix (the row order is $(1,1),(1,2),(2,1),(2,2)$):
\begin{eqnarray}
    M &=& \left[\begin{matrix}
            (\pi - \theta^1_1) \cos\theta^1_1 + \sin\theta^1_1\cos\theta^1_1 & (\pi - \theta^2_1) \cos\theta^2_1 + \sin\theta^2_1\cos\theta^1_1 \\ 
            (\pi - \theta^1_1) \cos\theta^1_2 + \sin\theta^1_1\cos\theta^1_2 & (\pi - \theta^2_1) \cos\theta^2_2 + \sin\theta^2_1\cos\theta^1_2 \\ 
            (\pi - \theta^1_2) \cos\theta^1_1 + \sin\theta^1_2\cos\theta^2_1 & (\pi - \theta^2_2) \cos\theta^2_1 + \sin\theta^2_2\cos\theta^2_1 \\ 
            (\pi - \theta^1_2) \cos\theta^1_2 + \sin\theta^1_2\cos\theta^2_2 & (\pi - \theta^2_2) \cos\theta^2_2 + \sin\theta^2_2\cos\theta^2_2
    \end{matrix}\right] \\
    &=& 
         \left[\begin{matrix}
            \pi & (\pi - \theta)\cos\theta + \sin\theta \\
            \pi\cos\theta & (\pi - \theta) + \sin\theta\cos\theta \\
            (\pi - \theta) + \sin\theta\cos\theta & \pi\cos\theta \\
            (\pi- \theta)\cos\theta + \sin\theta & \pi
        \end{matrix}\right]
\end{eqnarray}
since $\theta_1^2 = \theta_2^1 = \theta$, $\theta^1_1=\theta^2_2=0$. Similarly we could write $M\opt$: 
\begin{equation}
    M\opt = \left[\begin{matrix}
            (\pi - \theta^{*1}_1)\cos\theta^{*1}_1 + \sin\theta^{*1}_1\cos\theta^1_1 & (\pi - \theta^{*2}_1)\cos\theta^{*2}_1 + \sin\theta^{*2}_1\cos\theta^1_1 \\
            (\pi - \theta^{*1}_1)\cos\theta^{*1}_2 + \sin\theta^{*1}_1\cos\theta^1_2 & (\pi - \theta^{*2}_1)\cos\theta^{*2}_2 + \sin\theta^{*2}_1\cos\theta^1_2 \\
            (\pi - \theta^{*1}_2)\cos\theta^{*1}_1 + \sin\theta^{*1}_2\cos\theta^2_1 & (\pi - \theta^{*2}_2)\cos\theta^{*2}_1 + \sin\theta^{*2}_2\cos\theta^2_1 \\
            (\pi - \theta^{*1}_2)\cos\theta^{*1}_2 + \sin\theta^{*1}_2\cos\theta^2_2 & (\pi - \theta^{*2}_2)\cos\theta^{*2}_2 + \sin\theta^{*2}_2\cos\theta^2_2
        \end{matrix}\right]
\end{equation}
In this case, 
\begin{equation}
    M_r = \left[\begin{matrix}
            h(\theta^1_1) & h(\theta^2_1) \\
            h(\theta^1_2) & h(\theta^2_2)
        \end{matrix}\right],\quad\quad
    M\opt_r = \left[\begin{matrix}
            h(\theta^{*1}_1) & h(\theta^{*2}_1) \\
            h(\theta^{*1}_2) & h(\theta^{*2}_2)
        \end{matrix}\right]
\end{equation}
Therefore, if we know $\theta_1^2 = \theta_2^1$, $\theta_1^{*1}$, $\theta_1^{*2}$, $\theta_2^{*1}$ and $\theta_2^{*2}$, then we could compute $M$ and $M\opt$ and solve a linear equation to get the magnitude of $\vw_1$ and $\vw_2$, which collectly identify the critical points. Note that $M$ is a $4$-by-$2$ matrix, so critical point only happens if the matrix has singular structure.

\textbf{Global Optimum.} One special case is when $\theta_1^2 = \theta_2^1 = \theta_1^{*2} = \theta_2^{*1} = \pi / 2$ and $\theta_1^{*1} = \theta_2^{*2} = 0$, in this case, we have:
\begin{equation}
    M = M\opt = \left[\begin{matrix}
            \pi & 1 \\
            0 & \pi / 2\\
            \pi / 2 & 0\\
            1 & \pi
        \end{matrix}\right]
\end{equation}
and thus $\|\vw_j\| = \|\vw\opt_j\|$ is the unique solution. 

When $K = 2$, the following conjecture is empirically correct.
\begin{conjecture}
    If $\ve\opt$ is in the interior of $\mathrm{Cone}(\ve_1, \ve_2)$, then $L_{12}(\theta\opt_1, \theta\opt_2, \theta^1_2) > 0$. If $\ve\opt$ is in the exterior, then $L_{12} < 0$. If $\ve\opt$ is on the boundary then $L_{12} = 0$. Same for $L_{21}$.\label{conj:2d}
\end{conjecture}
\begin{remark}
    Note that $L_{1j}$ can be written as the following:
    \begin{eqnarray}
        L_{1j}(\ve\opt, \{\ve_1, \ve_2\}) &=& m_{1j}\opt - \left[h(\theta\opt_1), h(\theta\opt_2)\right] M_r^{-1}\vm_{1j} \\
        &=&\left[(\pi - \theta\opt_1)\ve\opt + \sin\theta_1\opt\ve_1\right]\trans \ve_j \\
        &-& \left[\alpha(\theta_1\opt, \theta_2\opt, \theta), \beta(\theta_1\opt, \theta_2\opt, \theta)\right]\left[
        \begin{matrix}
            (\pi - \theta^1_1)\ve\trans_1 + \sin\theta^1_1\ve\trans_1 \\
            (\pi - \theta^2_1)\ve\trans_2 + \sin\theta^2_1\ve\trans_1 
        \end{matrix}
    \right] \ve_j
    \end{eqnarray}
    Here we have 
    \begin{equation}
        [\alpha, \beta] = \left[\alpha(\theta_1\opt, \theta_2\opt, \theta), \beta(\theta_1\opt, \theta_2\opt, \theta)\right] \equiv [h(\theta\opt_1), h(\theta\opt_2)] M_r^{-1}
    \end{equation}
    We know that $L_{11} = 0$ by Proposition~\ref{proposition:property-L}. Therefore
    \begin{eqnarray}
        \vu_{1j} &\equiv& (\pi - \theta\opt_1)\ve\opt + \sin\theta_1\opt\ve_1 - \left[(\pi - \theta^1_1)\ve_1 + \sin\theta^1_1\ve_1, (\pi - \theta^2_1)\ve_2 + \sin\theta^2_1\ve_1\right]
        \left[
        \begin{matrix}
            \alpha(\theta_1\opt, \theta_2\opt, \theta^1_2) \\
            \beta(\theta_1\opt, \theta_2\opt, \theta^1_2)
        \end{matrix}
        \right] \nonumber \\
    &=& (\pi - \theta\opt_1)\ve\opt + \sin\theta_1\opt\ve_1 - \left[\pi\ve_1, (\pi - \theta)\ve_2 + \sin\theta\ve_1\right]
        \left[
        \begin{matrix}
            \alpha(\theta_1\opt, \theta_2\opt, \theta) \\
            \beta(\theta_1\opt, \theta_2\opt, \theta)
        \end{matrix}
    \right]
    \end{eqnarray}
    is perpendicular to $\ve_1$. So if we compute the inner product between $\vu_{12}$ and $\ve^\perp_1$ (the unit vector that is in $\Pi_*$ and is orthogonal to $\ve_1$), we get
    \begin{equation}
        \vu_{12}\trans\ve^\perp_1 = (\pi - \theta\opt_1)\sin\theta\opt_1 - \left[(\pi-\theta)\sin\theta\right]\beta \label{eq:prove-gt-zero}
    \end{equation}
    Since $\ve_2 = \cos\theta \ve_1 + \sin\theta\ve_1^\perp$ so we have:
    \begin{equation}
        L_{12}(\ve\opt, \{\ve_1, \ve_2\}) = \vu_{12}\trans \ve_2 = \sin\theta (\vu_{12}\trans\ve^\perp_1)
    \end{equation}
    Note that Eqn.~\ref{eq:prove-gt-zero} is a function with 2-variables $\theta$ and $\theta\opt_1$ ($\theta\opt_2$ is determined by $\theta$ and $\theta\opt_1$, depending on whether $\ve\opt$ is inside or outside $Cone(\ve_1, \ve_2)$). And we could verify it numerically. 
\end{remark}

\begin{theorem}
    If Conjecture~\ref{conj:2d} is correct, then for 2 ReLU network, $(\vw_1,\vw_2)$ ($\vw_1 \neq \vw_2$) is not a critical point, if they both are in $\mathrm{Cone}(\vw\opt_1, \vw\opt_2)$, or both out of it.
\end{theorem}
\begin{proof}
    If both $\vw\opt_1$ and $\vw\opt_2$ are inside $\mathrm{Cone}(\vw_1, \vw_2)$, then from Conjecture~\ref{conj:2d}, we have 
    \begin{equation}
        L_{12}(\theta^{k*}_1, \theta^{k*}_2, \theta^1_2) > 0
    \end{equation}
    for $k = 1, 2$. Since $K=2$ we could simply apply Thm.~\ref{lemma:necessary-condition} to say $(\vw_1, \vw_2)$ is not a critical point. Similary we prove the case for both $\vw\opt_1$ and $\vw\opt_2$ outside $\mathrm{Cone}(\vw_1, \vw_2)$.\qed
\end{proof}

\section{Convergence Analysis}
\subsection{Single ReLU case}
In this subsection, we mainly deal with the following dynamics:
\begin{equation}
    \ee{\nabla_\vw J} = \frac{N}{2} (\vw - \vw\opt) + \frac{N}{2\pi}\left(\theta\vw\opt - \frac{\|\vw\opt\|}{\|\vw\|}\sin\theta\vw \right) \label{eq:dynamics}
\end{equation}

\begin{theorem}
    In the region $\|\vw\tone - \vw\opt\| < \|\vw\opt\|$, following the dynamics (Eqn.~\ref{eq:dynamics}), the Lyapunov function $V(\vw) = \frac{1}{2}\|\vw - \vw\opt\|^2$ has $\dot V < 0$ and the system is asymptotically stable and thus $\vw\t\rightarrow\vw\opt$ when $t\rightarrow +\infty$.
\end{theorem}
\begin{proof}
    Denote that $\Omega = \{\vw: \|\vw\tone - \vw\opt\| < \|\vw\opt\|\}$. Note that 
\begin{equation}
    \dot V = -(\vw - \vw\opt)\trans \nabla_{\vw} J = -\vy\trans M \vy
\end{equation}
where $\vy = [\lenwopt, \lenw]\trans$ and $M$ is the following $2$-by-$2$ matrix:
\begin{equation}
    M = \frac12 \begin{bmatrix}
        \sin2\theta + 2\pi - 2\theta & -(2\pi - \theta)\cos\theta - \sin\theta \\
        -(2\pi - \theta)\cos\theta - \sin\theta & 2\pi
    \end{bmatrix}
\end{equation}
In the following we will show that $M$ is positive definite when $\theta \in (0, \pi/2]$. It suffices to show that $M_{11} > 0$, $M_{22} > 0$ and $\mathrm{det}(M)> 0$. The first two are trivial, while the last one is:
\begin{eqnarray}
    4\mathrm{det}(M) &=&  2\pi(\sin2\theta + 2\pi-2\theta) - \left[(2\pi - \theta)\cos\theta + \sin\theta\right]^2 \\
    &=& 2\pi(\sin2\theta + 2\pi-2\theta) - \left[(2\pi - \theta)^2\cos^2\theta + (2\pi -\theta)\sin2\theta +\sin^2\theta\right] \\
    &=& (4\pi^2 - 1)\sin^2\theta - 4\pi\theta + 4\pi\theta\cos^2\theta - \theta^2\cos^2\theta + \theta\sin2\theta \\
    &=& (4\pi^2 - 4\pi\theta - 1)\sin^2\theta + \theta\cos\theta (2\sin\theta - \theta\cos\theta)
\end{eqnarray}
Note that $4\pi^2 - 4\pi\theta - 1 = 4\pi(\pi- \theta) - 1 > 0$ for $\theta\in[0, \pi/2]$, and $g(\theta) = \sin\theta - \theta\cos\theta \ge 0$ for $\theta\in[0, \pi/2]$ since $g(0) = 0$ and $g'(\theta) \ge 0$ in this region. Therefore, when $\theta\in(0, \pi/2]$, $M$ is positive definite. 

When $\theta = 0$, $M(\theta) = \pi [ 1, -1; -1, 1]$ and is semi-positive definite, with the null eigenvector being $\frac{\sqrt{2}}{2} [1, 1]$, i.e., $\|\vw\| = \|\vw\opt\|$. However, along $\theta = 0$, the only $\vw$ that satisfies $\|\vw\| = \|\vw\opt\|$ is $\vw= \vw\opt$. Therefore, $\dot V = -\vy\trans M\vy < 0$ in $\Omega$. 
Note that although this region could be expanded to the entire open half-space $\mathcal{H} = \{\vw: \vw\trans\vw\opt > 0\}$, it is not straightforward to prove the convergence in $\mathcal{H}$, since the trajectory might go outside $\mathcal{H}$. On the other hand, $\Omega$ is the level set $V < \frac{1}{2}\|\vw\opt\|^2$ so the trajectory starting within $\Omega$ remains inside.
\qed
\end{proof}

\begin{figure}[h!]
    \centering
    \includegraphics[width=0.7\textwidth]{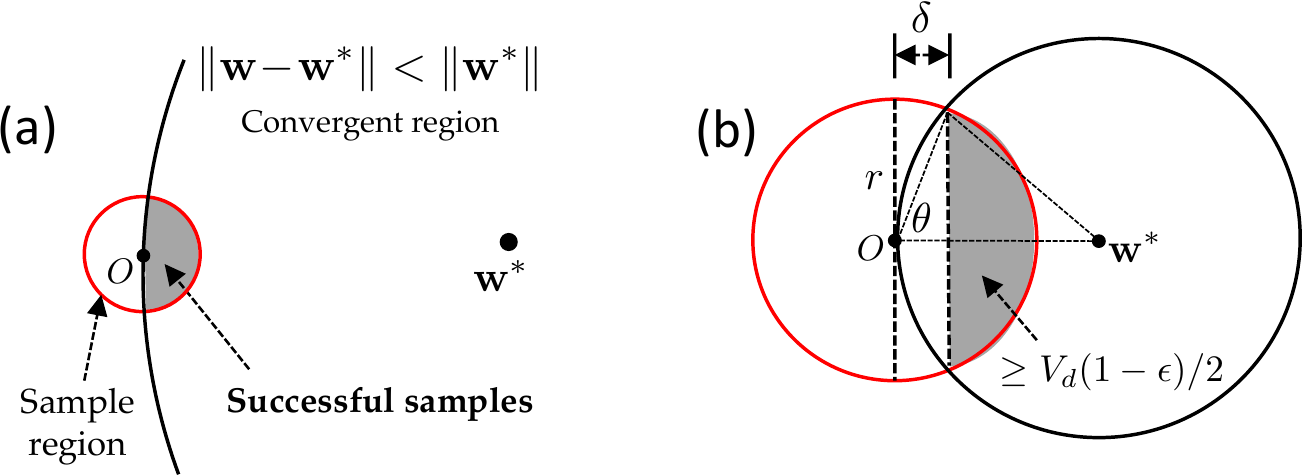}
    \caption{\textbf{(a)} Sampling strategy to maximize the probability of convergence. \textbf{(b)} Relationship between sampling range $r$ and desired probability of success $(1 - \epsilon)/2$.} \label{fig:convergence-analysis}
\end{figure}

\begin{theorem}
    When $K = 1$, the dynamics in Eqn.~\ref{eq:two-layer-fix-top} converges to $\vw\opt$ with probability at least $(1 - \epsilon)/2$, if the initial value $\vw\tone$ is sampled uniformly from $B_r = \{\vw : \|\vw\|\le r\}$ with:
    \begin{equation}
        r \le \epsilon\sqrt{\frac{2\pi}{d + 1}}\|\vw\opt\| 
    \end{equation}
\end{theorem}
\begin{proof}
    Given a ball of radius $r$, we first compute the ``gap'' $\delta$ of sphere cap (Fig.~\ref{fig:convergence-analysis}(b)). First $\cos\theta = \frac{r}{2\|\vw\opt\|}$, so $\delta = r\cos\theta = \frac{r^2}{2\|\vw\opt\|}$. Then a sufficient condition for the probability argument to hold, is to ensure that the volume $V_\mathrm{shaded}$ of the shaded area is greater than $\frac{1 - \epsilon}{2}V_d(r)$, where $V_d(r)$ is the volume of $d$-dimensional ball of radius $r$. Since $V_\mathrm{shaded} \ge \frac{1}{2}V_d(r) - \delta V_{d-1}$, it suffices to have:
\begin{equation}
    \frac{1}{2}V_d(r) - \delta V_{d-1} \ge \frac{1 - \epsilon}{2}V_d(r)
\end{equation}
which gives 
\begin{equation}
    \delta \le \frac{\epsilon}{2}\frac{V_d}{V_{d-1}}
\end{equation}
Using $\delta = \frac{r^2}{2\|\vw\opt\|}$ and $V_d(r) = V_d(1) r^d$, we thus have: 
\begin{equation}
    r \le \epsilon\frac{V_d(1)}{V_{d-1}(1)}\|\vw\opt\| 
\end{equation}
where $V_d(1)$ is the volume of the unit ball. Since the volume of $d$-dimensional unit ball is 
\begin{equation}
    V_d(1) = \frac{\pi^{d/2}}{\Gamma(d/2 + 1)}
\end{equation}
where $\Gamma(x) = \int_0^\infty t^{x-1} e^{-t}\dd t$. So we have
\begin{equation}
    \frac{V_d(1)}{V_{d-1}(1)} = \sqrt{\pi}\frac{\Gamma(d/2 + 1/2)}{\Gamma(d/2 + 1)}
\end{equation}
From Gautschi's Inequality
\begin{equation}
    x^{1-s} < \frac{\Gamma(x+1)}{\Gamma(x+s)} < (x+s)^{1-s} \quad\quad x > 0, 0 < s < 1
\end{equation}
with $s = 1/2$ and $x = d/2$ we have:
\begin{equation}
    \left(\frac{d + 1}{2}\right)^{-1/2} < \frac{\Gamma(d/2 + 1/2)}{\Gamma(d/2 + 1)} < \left(\frac{d}{2}\right)^{-1/2}
\end{equation}
Therefore, it suffices to have 
\begin{equation}
    r \le \epsilon\sqrt{\frac{2\pi}{d + 1}}\|\vw\opt\| \label{eq:sampling-criterion-appendix}
\end{equation}
Note that this upper bound is tight when $\delta\rightarrow 0$ and $d\rightarrow +\infty$, since all inequality involved asymptotically becomes equal.
\qed
\end{proof}

\subsection{Multiple ReLU case}
\textbf{Explanation of Eqn. 18}. We first write down the dynamics to be studied:
\begin{equation}
    -\ee{\nabla_{\vw_j} J} = \sum_{j'=1}^K \ee{F(\ve_j, \vw\opt_{j'})} - \ee{F(\ve_j, \vw_{j'})} \label{eq:two-layer-fix-top} 
\end{equation}
We first define $f(\vw_j, \vw_{j'}, \vw\opt_{j'}) = F(\vw_j/\|\vw_j\|, \vw\opt_{j'}) - F(\vw_j/\|\vw_j\|, \vw_{j'})$. Therefore, the dynamics can be written as:
\begin{equation}
    -\ee{\nabla_{\vw_j} J} = \sum_{j'} \ee{f(\vw_j, \vw_{j'}, \vw\opt_{j'})} 
\end{equation}
Suppose we have a finite group $\mathcal{P} = \{P_j\}$ which is a subgroup of orthognoal group $O(d)$. $P_1$ is the identity element. If $\vw$ and $\vw\opt$ have the following symmetry: $\vw_{j} = P_j\vw$ and $\vw\opt_j = P_j\vw\opt$, then RHS of Eqn.~\ref{eq:two-layer-fix-top} can be simplified as follows:

\begin{eqnarray}
    -\ee{\nabla_{\vw_j} J} &=& \sum_{j'} \ee{f(\vw_j, \vw_{j'}, \vw\opt_{j'})} = \sum_{j'} \ee{f(P_j\vw, P_{j'}\vw, P_{j'}\vw\opt)} \nonumber \\
    &=& \sum_{j''} \ee{f(P_j\vw, P_jP_{j''}\vw, P_jP_{j''}\vw\opt)} \quad (\{P_j\}_{j=1}^K\ \mathrm{is\ a\ group}) \nonumber \\
    &=& P_j \sum_{j''} \ee{f(\vw, P_{j''}\vw, P_{j''}\vw\opt)} \quad (\|P\vw_1\|=\|\vw_1\|,\ \angle(P\vw_1, P\vw_2) = \angle(\vw_1, \vw_2)) \nonumber \\
    &=& -P_j \ee{\nabla_{\vw_1} J} 
\end{eqnarray}
which means that if all $\vw_j$ and $\vw\opt_j$ are symmetric under the action of cyclic group, so does their expected gradient. Therefore, the trajectory $\{\vw\t\}$ keeps such cyclic structure. Instead of solving a system of $K$ equations, we only need to solve one:
\begin{equation}
    -\ee{\nabla_{\vw} J} = \sum_{j=1}^K \ee{f(\vw, P_j\vw, P_j\vw\opt)} \label{eq:symmetric-one} 
\end{equation}

\begin{theorem}
\label{thm:symmetric-breaking}
For a bias-free two-layered ReLU network
\begin{equation}
    g(\vx; \vw) = \sum_j \sigma(\vw_j\trans\vx)
\end{equation}
that takes spherical Gaussian inputs, if the teacher's parameters $\{\vw\opt_j\}$ form a set of orthnomal bases, then:
\begin{itemize}
    \item[(1)] When the student parameters is initialized to be $[x\tone, y\tone, \ldots, y\tone]$ under the basis of $\vw\opt$, where $(x\tone, y\tone)\in \Omega = \{x\in(0,1], y\in[0,1], x>y\}$, then the dynamics (Eqn.~\ref{eq:two-layer-fix-top}) converges to teacher's parameters $\{\vw\opt_j\}$ (or $(x,y)=(1,0)$); 
    \item[(2)] when $x\tone = y\tone \in (0, 1]$, then it converges to a saddle point $x = y = \frac{1}{\pi K}(\sqrt{K-1} - \arccos(1/\sqrt{K}) + \pi)$. 
\end{itemize}
\end{theorem}
This theorem is quite complicated. We will start with a few lemmas and gradually come to the conclusion.

First, if $\vw\tone = [x, y, y, \ldots, y]$ under the bases $\{\vw\opt_j\}_{j=1}^K$, then from simple computation we know that $\vw\t$ also follows this pattern. Therefore, we only need to study the following 2D dynamics related to $x$ and $y$:
\begin{eqnarray}
    -\frac{2\pi}{N}
    \mathbb{E}
    \begin{bmatrix}
        \nabla_x J \\ 
        \nabla_y J
    \end{bmatrix}
    &=& -\left\{ [(\pi - \phi)(x - 1 + (K-1)y)] 
    \begin{bmatrix} 
        1 \\  
        1
    \end{bmatrix}
    + \begin{bmatrix}
        \theta \\
        \phi\opt - \phi
    \end{bmatrix}
    + \phi\begin{bmatrix}
        x - 1 \\
        y
    \end{bmatrix}
\right\} \nonumber \\
    &+& [(K-1)(\alpha\sin\phi\opt - \sin\phi) + \alpha\sin\theta]
    \begin{bmatrix}
        x\\
        y
    \end{bmatrix} \label{eq:2d-dynamics}
\end{eqnarray}
Here the symmetrical factor $(\alpha \equiv \lenwoptj{j'} / \lenwj{j}, \theta \equiv \theta_j^{*j}, \phi \equiv \theta_j^{j'}, \phi\opt \equiv \theta_j^{*j'})$ are defined as follows:
\begin{equation}
    \alpha = (x^2 + (K-1)y^2)^{-1/2},\quad \cos\theta = \alpha x,\quad \cos\phi\opt = \alpha y,\quad \cos\phi = \alpha^2(2xy + (K-2)y^2) \label{eq:angles-two-layers}
\end{equation}
Now we start a sequence of lemmas.

\begin{lemma}
    \label{lemma:prepositions}
    For $\phi\opt$, $\theta$ and $\phi$ defined in Eqn.~\ref{eq:angles-two-layers}:
    \begin{eqnarray}
        \alpha &\equiv& (x^2 + (K-1)y^2)^{-1/2} \\
        \cos\theta &\equiv& \alpha x \\
        \cos\phi\opt &\equiv& \alpha y \\
        \cos\phi &\equiv& \alpha^2(2xy + (K-2)y^2)
    \end{eqnarray}
    we have the following relations in the triangular region $\Omega_{\epsilon_0} = \{ (x, y): x \ge 0, y \ge 0, x \ge y + \epsilon_0\}$ (Fig.~\ref{fig:f}(c)):
    \begin{itemize}
        \item[(1)] $\phi, \phi\opt\in [0, \pi/2]$ and $\theta \in [0, \theta_0)$ where $\theta_0 = \arccos\frac{1}{\sqrt{K}}$.
        \item[(2)] $\cos\phi = 1 - \alpha^2 (x-y)^2$ and $\sin\phi = \alpha(x-y)\sqrt{2 - \alpha^2(x-y)^2}$.
        \item[(3)] $\phi\opt \ge \phi$ (equality holds only when $y = 0$) and $\phi\opt > \theta$.
    \end{itemize}
\end{lemma}
\begin{proof}
    Propositions (1) and (2) are computed by direct calculations. In particular, note that since $\cos\theta = \alpha x = 1 / \sqrt{1 + (K-1)(y/x)^2}$ and $x > y \ge 0$, we have $\cos\theta \in (1/\sqrt{K}, 1]$ and $\theta\in [0, \theta_0)$. For Preposition (3), $\phi\opt = \arccos \alpha y > \theta = \arccos \alpha x$ because $x> y$. Finally, for $x > y > 0$, we have
    \begin{equation}
        \frac{\cos\phi}{\cos\phi\opt} = \frac{\alpha^2(2xy + (K-2)y^2)}{\alpha y} = \alpha (2x + (K-2)y) > \alpha(x + (K-1)y) > 1
    \end{equation}
    The final inequality is because $K \ge 2$, $x, y > 0$ and thus $(x+(K-1)y)^2 > x^2+(K-1)^2y^2 > x^2 + (K-1)y^2 = \alpha^{-2}$. Therefore $\phi\opt > \phi$. If $y = 0$ then $\phi\opt = \phi$.\qed
\end{proof}

\begin{figure}
    \centering
    \includegraphics[width=0.3\textwidth]{2D-convergence-crop.pdf}
    \caption{The region $\Omega_\epsilon$ considered in the analysis of Eqn.~\ref{eq:2d-dynamics}.} \label{fig:region}
\end{figure}

\begin{lemma}
    For the dynamics defined in Eqn.~\ref{eq:2d-dynamics}, there exists $\epsilon_0 > 0$ so that the trianglar region $\Omega_{\epsilon_0} = \{ (x, y): x \ge 0, y \ge 0, x \ge y + \epsilon_0\}$ (Fig.~\ref{fig:region}) is a convergent region. That is, the flow goes inwards for all three edges and any trajectory starting in $\Omega_{\epsilon_0}$ stays. 
\label{lemma:convergent-region}
\end{lemma}
\begin{proof}
We discuss the three boundaries as follows:

\textbf{Case 1: $y = 0, 0 \le x \le 1$, horizontal line}. In this case, $\theta = 0$, $\phi = \pi / 2$ and $\phi\opt = \pi/2$. The $y$ component of the dynamics in this line is:
\begin{equation}
    f_1 \equiv -\frac{2\pi}{N} \nabla_y J = -\frac{\pi}{2}(x-1) \ge 0
\end{equation}
So $-\nabla_y J$ points to the interior of $\Omega$.

\textbf{Case 2: $x = 1, 0 \le y \le 1$, vertical line}. In this case, $\alpha \le 1$ and the $x$ component of the dynamics is:
\begin{eqnarray}
    f_2 \equiv -\frac{2\pi}{N} \nabla_x J &=& -(\pi - \phi)(K-1)y - \theta + (K-1)(\alpha\sin\phi\opt - \sin\phi) + \alpha\sin\theta \\
    &=& -(K-1)\left[(\pi - \phi)y - (\alpha\sin\phi\opt - \sin\phi)\right] + (\alpha\sin\theta - \theta)
\end{eqnarray}
Note that since $\alpha \le 1$, $\alpha\sin\theta \le \sin\theta \le \theta$, so the second term is non-positive. For the first term, we only need to check whether $(\pi - \phi)y - (\alpha\sin\phi\opt - \sin\phi)$ is nonnegative. Note that 
\begin{eqnarray}
    & & (\pi - \phi)y - (\alpha\sin\phi\opt - \sin\phi) \\
    &=& (\pi - \phi)y + \alpha (x-y)\sqrt{2 - \alpha^2(x-y)^2} - \alpha\sqrt{1 - \alpha^2y^2} \\
    &=& y\left[\pi - \phi - \alpha\sqrt{2 - \alpha^2(x-y)^2}\right] + \alpha\left[x\sqrt{2 - \alpha^2(x-y)^2} - \sqrt{1-\alpha^2y^2}\right]
\end{eqnarray}
In $\Omega$ we have $(x-y)^2 \le 1$, combined with $\alpha \le 1$, we have $1 \le \sqrt{2-\alpha^2(x-y)^2} \le \sqrt{2}$ and $\sqrt{1 - \alpha^2y^2} \le 1$. Since $x = 1$, the second term is nonnegative. For the first term, since $\alpha \le 1$, 
\begin{equation}
    \pi - \phi - \alpha\sqrt{2-\alpha^2(x-y)^2} \ge \pi - \frac{\pi}{2} - \sqrt{2} > 0
\end{equation}
So $(\pi - \phi)y - (\alpha\sin\phi\opt - \sin\phi) \ge 0$ and $-\nabla_x J \le 0$, pointing inwards.

\textbf{Case 3: $x = y + \epsilon$, $0\le y\le 1$, diagonal line}. We compute the inner product between $(-\nabla_x J, -\nabla_y J)$ and $(1, -1)$, the inward normal of $\Omega$ at the line. Using $\phi \le \frac{\pi}{2}\sin\phi$ for $\phi \in [0, \pi/2]$ and $\phi\opt - \theta = \arccos\alpha y - \arccos\alpha x  \ge 0$ when $x \ge y$, we have:
\begin{eqnarray}
    f_3(y, \epsilon) \equiv  
    -\frac{2\pi}{N} 
    \begin{bmatrix}
        \nabla_x J \\
        \nabla_y J
    \end{bmatrix}\trans
    \begin{bmatrix}
        1 \\
        -1
    \end{bmatrix}
    &=& \phi\opt - \theta - \epsilon \phi + \left[(K-1)(\alpha\sin\phi\opt - \sin\phi) + \alpha\sin\theta\right]\epsilon \label{eq:f3} \\
    &\ge& \epsilon(K-1)\left[\alpha\sin\phi\opt - \left(1 + \frac{\pi}{2(K-1)}\right)\sin\phi\right] \nonumber \\
        &=& \epsilon\alpha(K-1)\left[\sqrt{1 - \alpha^2y^2} - \epsilon\left(1 + \frac{\pi}{2(K-1)}\right)\sqrt{2-\alpha^2\epsilon^2}\right] \nonumber 
\end{eqnarray}
Note that for $y > 0$:
\begin{equation}
    \alpha y = \frac{1}{\sqrt{ (x/y)^2 + (K-1)}} = \frac{1}{\sqrt{ (1 + \epsilon/y)^2 + (K-1)}} \le \frac{1}{\sqrt{K}}
\end{equation}
For $y = 0$, $\alpha y = 0 < \sqrt{1/K}$. So we have $\sqrt{1 - \alpha^2y^2} \ge \sqrt{1 - 1/K}$. And $\sqrt{2 - \alpha^2\epsilon^2} \le \sqrt{2}$. Therefore $f_3 \ge \epsilon\alpha(K-1)(C_1 - \epsilon C_2)$ with $C_1 \equiv \sqrt{1 - 1/K} > 0$ and $C_2 \equiv \sqrt{2}(1 + \pi/2(K-1)) > 0$. With $\epsilon = \epsilon_0 > 0$ sufficiently small, $f_3 > 0$.
\qed
\end{proof}

\begin{lemma}[Reparametrization]
    Denote $\epsilon = x - y > 0$. The terms $\alpha x$, $\alpha y$ and $\alpha \epsilon$ involved in the trigometric functions in Eqn.~\ref{eq:2d-dynamics} has the following parameterization:
    \begin{eqnarray}
        \alpha
        \begin{bmatrix}
            y \\
            x \\
            \epsilon
        \end{bmatrix}
        = \frac{1}{K}\begin{bmatrix}
            \beta - \beta_2 \\ 
            \beta + (K-1) \beta_2\\ 
            K\beta_2
        \end{bmatrix} \label{eq:beta-parametrization}
    \end{eqnarray}
    where $\beta_2 = \sqrt{ (K - \beta^2) / (K-1)}$. The reverse transformation is given by $\beta = \sqrt{K - (K-1)\alpha^2\epsilon^2}$. Here $\beta \in [1, \sqrt{K})$ and $\beta_2 \in (0, 1]$. In particular, the critical point $(x, y) = (1, 0)$ corresponds to $(\beta, \epsilon) = (1, 1)$. As a result, all trigometric functions in Eqn.~\ref{eq:2d-dynamics} only depend on the single variable $\beta$. In particular, the following relationship is useful:
    \begin{equation}
        \beta = \cos\theta + \sqrt{K-1}\sin\theta \label{eq:beta-theta}
    \end{equation}
\end{lemma}
\begin{proof}
    This transformation can be checked by simple algebraic manipulation. For example:
    \begin{equation}
        \frac{1}{\alpha K}(\beta - \beta_2) = \frac{1}{K}\left(\sqrt{\frac{K}{\alpha^2} - (K-1)\epsilon^2} - \epsilon\right) = \frac{1}{K}\left(\sqrt{(Ky+\epsilon)^2} - \epsilon\right) = y
    \end{equation}
    To prove Eqn.~\ref{eq:beta-theta}, first we notice that $K\cos\theta = K\alpha x = \beta+(K-1)\beta_2$. Therefore, we have $(K\cos\theta - \beta)^2 - (K-1)^2\beta_2^2 = 0$, which gives $\beta^2 - 2\beta\cos\theta + 1 - K\sin^2\theta = 0$. Solving this quadratic equation and notice that $\beta \ge 1$, $\theta \in [0, \pi/2]$ and we get:
    \begin{equation}
        \beta = \cos\theta + \sqrt{\cos^2\theta + K\sin^2\theta - 1} = \cos\theta + \sqrt{K - 1}\sin\theta
    \end{equation}
    \qed
\end{proof}

\begin{lemma}    
    After reparametrization (Eqn.~\ref{eq:beta-parametrization}), $f_3(\beta, \epsilon) \ge 0$ for $\epsilon \in (0, \beta_2/\beta]$. Furthermore, the equality is true only if $(\beta, \epsilon) = (1, 1)$ or $(y, \epsilon) = (0, 1)$.
    \label{lemma:single-critical-point}
\end{lemma}
\begin{proof}
    Applying the parametrization (Eqn.~\ref{eq:beta-parametrization}) to Eqn.~\ref{eq:f3} and notice that $\alpha\epsilon = \beta_2 = \beta_2(\beta)$, we could write
    \begin{equation}
        f_3 = h_1(\beta) - (\phi + (K-1)\sin\phi)\epsilon
    \end{equation}
    When $\beta$ is fixed, $f_3$ now is a monotonously decreasing function with respect to $\epsilon > 0$. Therefore, $f_3(\beta, \epsilon) \ge f_3(\beta, \epsilon')$ for $0 < \epsilon \le \epsilon' \equiv \beta_2/\beta$. If we could prove $f_3(\beta, \epsilon') \ge 0$ and only attain zero at known critical point $(\beta, \epsilon) = (1, 1)$, the proof is complete. 
    
    Denote $f_3(\beta, \epsilon') = f_{31} + f_{32}$ where
    \begin{eqnarray}
        f_{31}(\beta, \epsilon') &=& \phi\opt - \theta - \epsilon'\phi + \epsilon'\alpha\sin\theta\\
        f_{32}(\beta, \epsilon') &=& (K-1)(\alpha\sin\phi\opt - \sin\phi)\epsilon' 
    \end{eqnarray}
For $f_{32}$ it suffices to prove that $\epsilon'(\alpha\sin\phi\opt - \sin\phi) = \beta_2\sin\phi\opt - \frac{\beta_2}{\beta}\sin\phi \ge 0$, which is equivalent to $\sin\phi\opt - \sin\phi / \beta \ge 0$. But this is trivially true since $\phi\opt \ge \phi$ and $\beta \ge 1$. Therefore, $f_{32} \ge 0$. Note that the equality only holds when $\phi\opt = \phi$ and $\beta = 1$, which corresponds to the horizontal line $x\in (0, 1], y = 0$.

    For $f_{31}$, since $\phi\opt \ge \phi$, $\phi\opt > \theta$ and $\epsilon' \in (0, 1]$, we have the following:
    \begin{equation}
        f_{31} = \epsilon'(\phi\opt - \phi) + (1 - \epsilon')(\phi\opt - \theta) - \epsilon'\theta + \beta_2\sin\theta \ge - \epsilon'\theta + \beta_2\sin\theta \ge \beta_2\left(\sin\theta - \frac{\theta}{\beta}\right)
    \end{equation}
    And it reduces to showing whether $\beta\sin\theta - \theta$ is nonnegative. Using Eqn.~\ref{eq:beta-theta}, we have:
    \begin{equation}
        f_{33}(\theta) = \beta\sin\theta - \theta = \frac{1}{2}\sin2\theta + \sqrt{K-1}\sin^2\theta - \theta
    \end{equation}
    Note that $f'_{33} = \cos2\theta + \sqrt{K-1}\sin2\theta - 1 = \sqrt{K}\cos(2\theta - \theta_0) - 1$, where $\theta_0 = \arccos\frac{1}{\sqrt{K}}$. By Prepositions 1 in Lemma~\ref{lemma:prepositions}, $\theta\in [0, \theta_0)$. Therefore, $f'_{33} \ge 0$ and since $f_{33}(0) = 0$, $f_{33} \ge 0$. Again the equity holds when $\theta = 0$, $\phi\opt = \phi$ and $\epsilon' = 1$,  which is the critical point $(\beta, \epsilon) = (1, 1)$ or $(y, \epsilon) = (0, 1)$.
\qed
\end{proof}

\begin{lemma}
    For the dynamics defined in Eqn.~\ref{eq:2d-dynamics}, the only critical point ($\nabla_x J = 0$ and $\nabla_y J = 0$) within $\Omega_\epsilon$ is $(y, \epsilon) = (0, 1)$. 
    \label{lemma:single-critical-point}
\end{lemma}
\begin{proof}
    We prove by contradiction. Suppose $(\beta, \epsilon)$ is a critical point other than $\vw\opt$. A necessary condition for this to hold is $f_3 = 0$ (Eqn.~\ref{eq:f3}). By Lemma~\ref{lemma:single-critical-point}, $\epsilon > \epsilon' = \beta_2/\beta > 0$ and 
\begin{equation}
    \epsilon - 1 + Ky = \frac{1}{\alpha}\left(\beta_2 - \alpha + \beta - \beta_2\right) = \frac{\beta-\alpha}{\alpha} = \frac{\beta-\beta_2/\epsilon}{\alpha} > \frac{\beta-\beta_2/\epsilon'}{\alpha} = 0 
\end{equation}
So $\epsilon - 1 + Ky$ is strictly greater than zero. On the other hand, the condition $f_3 = 0$ implies that 
\begin{equation}
    \left((K-1)(\alpha\sin\phi\opt - \sin\phi) + \alpha\sin\theta\right) = -\frac{1}{\epsilon}(\phi\opt - \theta) + \phi  
\end{equation}
Using $\phi \in [0, \pi/2]$, $\phi\opt \ge \phi$ and $\phi\opt > \theta$, we have:
\begin{eqnarray}
    -\frac{2\pi}{N}\nabla_y J &=& -(\pi - \phi)(\epsilon - 1 + Ky) - (\phi\opt - \phi) - \phi y + \left((K-1)(\alpha\sin\phi\opt - \sin\phi) + \alpha\sin\theta\right) y \nonumber \\
    &=& -(\pi - \phi)(\epsilon - 1 + Ky) - (\phi\opt - \phi) - \frac{1}{\epsilon}(\phi\opt - \theta) y < 0
\end{eqnarray}
So the current point $(\beta, \epsilon)$ cannot be a critical point.\qed
\end{proof}

\begin{lemma}
    Any trajectory in $\Omega_{\epsilon_0}$ converges to $(y, \epsilon) = (1, 0)$, following the dynamics defined in Eqn.~\ref{eq:2d-dynamics}.
    \label{lemma:trajectory-convergence}
\end{lemma}
\begin{proof}
    We have Lyaponov function $V = \ee{E}$ so that $\dot V = -\ee{\nabla_{\vw} J \trans \nabla_{\vw} J} \le -\ee{\nabla_{\vw} J}\trans\ee{\nabla_{\vw} J} \le 0$. By Lemma~\ref{lemma:single-critical-point}, other than the optimal solution $\vw\opt$, there is no other symmetric critical point, $\nabla_{\vw} J \neq 0$ and thus $\dot V < 0$. On the other hand, by Lemma~\ref{lemma:convergent-region}, the triangular region $\Omega_{\epsilon_0}$ is convergent, in which the 2D dynamics is $C^\infty$ differentiable. Therefore, any 2D solution curve $\xi(t)$ will stay within. By Poincare$—$Bendixson theorem, when there is a unique critical point, the curve either converges to a limit circle or the critical point. However, limit cycle is not possible since $V$ is strictly monotonous decreasing along the curve. Therefore, $\xi(t)$ will converge to the unique critical point, which is $(y, \epsilon) = (1, 0)$ and so does the symmetric system (Eqn.~\ref{eq:two-layer-fix-top}).\qed
\end{proof}

\begin{lemma}
When $x = y \in (0, 1]$, the 2D dynamics (Eqn.~\ref{eq:2d-dynamics}) reduces to the following 1D case:
    \begin{equation}
        -\frac{2\pi}{N}\nabla_x J = -\pi K (x - x_*)
    \end{equation}
    where $x_* = \frac{1}{\pi K}(\sqrt{K-1} - \arccos(1/\sqrt{K}) + \pi)$. Furthermore, $x_*$ is a convergent critical point.
    \label{lemma:saddle}
\end{lemma}
\begin{proof}
    The 1D system can be computed with simple algebraic manipulations (note that when $x = y$, $\phi = 0$ and $\theta = \phi\opt = \arccos(1/\sqrt{K})$). Note that the 1D system is linear and its close form solution is $x\t = x_0 + C e^{-K/2N t}$ and thus convergent.\qed 
\end{proof}

Combining Lemma~\ref{lemma:trajectory-convergence} and Lemma~\ref{lemma:saddle} yields Thm.~\ref{thm:symmetric-breaking}.

\section{Simulations}
No theorems is provided.

\section{Extension to multilayer ReLU network}
\begin{proposition}
    For neural network with ReLU nonlinearity and using $l_2$ loss to match with a teacher network of the same size, the gradient inflow $\vg_j$ for node $j$ at layer $c$ has the following form:
\begin{equation}
    \vg_j = Q_j \sum_{j'} (Q_{j'}\vu_{j'} - Q\opt_{j'} \vu\opt_{j'})
\end{equation}
where $Q_{j}$ and $Q\opt_{j}$ are $N$-by-$N$ diagonal matrices. For any $k\in [c+1]$, $Q_k = \sum_{j\in [c]} w_{jk}D_jQ_j$ and similarly for $Q\opt_k$. The gradient with respect to $\vw_j$ (the parameters immediately under node $j$), is computed as:
\begin{equation}
    \nabla_{\vw_j} J = X_c^T D_j\trans \vg_j
\end{equation}
\end{proposition}
\begin{proof}
    We prove by induction on layer. For the first layer, there is only one node with $\vg = \vu - \vv$, therefore $Q_j = Q_{j'} = I$. Suppose the condition holds for all node $j\in [c]$. Then for node $k\in [c+1]$, we have:
\begin{eqnarray}
    \vg_k &=& \sum_j w_{jk} D_j \vg_j = \sum_j w_{jk} D_j Q_j\left(\sum_{j'} Q_{j'}\vu_{j'} - \sum_{j'} Q\opt_{j'}\vu\opt_{j'}\right)\nonumber \\
    &=& \sum_j w_{jk} D_j Q_j\left(\sum_{j'}Q_{j'}\sum_{k'} D_{j'} w_{jk'} \vu_{k'} - \sum_{j'} Q\opt_{j'}\sum_{k'} D\opt_{j'} w\opt_{jk'} \vu\opt_{k'}\right)\nonumber \\
    &=& \sum_{j} w_{jk} D_j Q_j \sum_{j'} Q_{j'} D_{j'} \sum_{k'} w_{jk'}\vu_{k'} - \sum_j w_{jk} D_j Q_j \sum_{j'} Q\opt_{j'} D\opt_{j'} \sum_{k'} w\opt_{jk'}\vu\opt_{k'} \nonumber \\
    &=& \sum_{k'}\left(\sum_{j} w_{jk} D_j Q_j\right)\left(\sum_{j'} Q_{j'} D_{j'} w_{jk'}\right) \vu_{k'} - \sum_{k'} \left(\sum_j w_{jk} D_j Q_j\right)\left(\sum_{j'} Q\opt_{j'} D\opt_{j'} w\opt_{jk'}\right)\vu\opt_{k'}\nonumber
\end{eqnarray}
Setting $Q_k = \sum_j w_{jk}D_jQ_j$ and $Q\opt_k = \sum_j w\opt_{jk}D\opt_jQ\opt_j$ (both are diagonal matrices), we thus have:
\begin{equation}
    \vg_{k} = \sum_{k'} Q_k Q_{k'} \vu_{k'} - Q_k Q\opt_{k'} \vu\opt_{k'} = Q_k \sum_{k'} Q_{k'} \vu_{k'}  - Q\opt_{k'} \vu\opt_{k'}
\end{equation} \qed
\end{proof}